\documentclass[a4paper,fleqn]{cas-sc}

\usepackage[authoryear,round]{natbib}
\usepackage{booktabs}
\usepackage{threeparttable}
\usepackage{rotating}
\usepackage{multirow}
\usepackage{amssymb} 
\usepackage{tocloft} 
\usepackage{caption} 
\usepackage{hyperref} 
\usepackage{indentfirst}
\usepackage{bbding}
\usepackage{amsthm}
\usepackage{bm}

\usepackage{graphicx} 
\usepackage{caption}  
\usepackage{tikz}
\usepackage{amsmath}
\usepackage{algorithm}
\usepackage{algorithmic}

\usepackage{tabularx}
\usepackage{amsmath}
\usepackage{multirow}
\usepackage{array} 
\usepackage{svg}

\usepackage{colortbl} 
\usepackage{xcolor}   
\usepackage{lineno}
\usepackage{makecell}
\usepackage{enumitem}
\usepackage{diagbox}
\usepackage{subcaption}
\usetikzlibrary{calc}

\def\tsc#1{\csdef{#1}{\textsc{\lowercase{#1}}\xspace}}
\tsc{WGM}
\tsc{QE}
\tsc{EP}
\tsc{PMS}
\tsc{BEC}
\tsc{DE}
\definecolor{mygreen}{RGB}{220,240,255} 
\definecolor{myblue}{RGB}{220,250,220}

\begin{document}
\let\WriteBookmarks\relax
\def\floatpagepagefraction{1}
\def\textpagefraction{.001}

\shorttitle{}



\title [mode = title]{Deep Reinforcement Learning for Drone Route Optimization in Post-Disaster Road Assessment}  


%
\author[1]{Huatian Gong}
\ead{huatian.gong@ntu.edu.sg}

\author[2]{Jiuh-Biing Sheu}
\cormark[1]
\ead{jbsheu@ntu.edu.tw}

\author[3]{Zheng Wang}
\ead{drwz@dlut.edu.cn}

\author[4]{Xiaoguang Yang}
\ead{yangxg@tongji.edu.cn}

\author[1]{Ran Yan}
\cormark[1]
\ead{ran.yan@ntu.edu.sg}

\affiliation[1]{organization={School of Civil and Environmental Engineering, Nanyang Technological University},
country={Singapore}}

\affiliation[2]{organization={Department of Business Administration, National Taiwan University},
country={Taiwan}}

\affiliation[3]{organization={School of Maritime Economics and Management, Dalian Maritime University},
country={China}}

\affiliation[4]{organization={The Key Laboratory of Road and Traffic Engineering of the Ministry of Education, Tongji University},
country={China}}

\cortext[cor1]{Corresponding author}










\begin{abstract}
Rapid post-disaster road damage assessment is critical for effective emergency response, yet traditional optimization methods suffer from excessive computational time and require domain knowledge for algorithm design, making them unsuitable for time-sensitive disaster scenarios. This study proposes an attention-based encoder-decoder model (AEDM) for rapid drone routing decision in post-disaster road damage assessment. The method employs deep reinforcement learning to determine high-quality drone assessment routes without requiring algorithmic design knowledge. A network transformation method is developed to convert link-based routing problems into equivalent node-based formulations, while a synthetic road network generation technique addresses the scarcity of large-scale training datasets. The model is trained using policy optimization with multiple optima (POMO) with multi-task learning capabilities to handle diverse parameter combinations. Experimental results demonstrate two key strengths of AEDM: it outperforms commercial solvers by 20--71\% and traditional heuristics by 23--35\% in solution quality, while achieving rapid inference (1--2 seconds) versus 100--2,000 seconds for traditional methods. The model exhibits strong generalization across varying problem scales, drone numbers, and time constraints, consistently outperforming baseline methods on unseen parameter distributions and real-world road networks. The proposed method effectively balances computational efficiency with solution quality, making it particularly suitable for time-critical disaster response applications where rapid decision-making is essential for saving lives. The source code for AEDM is publicly available at \url{https://github.com/PJ-HTU/AEDM-for-Post-disaster-road-assessment}.
\end{abstract} 




\maketitle

\section{Introduction}
\label{Introduction}

In recent years, the frequency and intensity of both natural disasters (e.g., earthquakes, hurricanes, and wildfires) and man-made crises (e.g., conflicts) have increased significantly, posing severe threats to human lives, property, and economic stability \citep{avishan2023humanitarian}. For example, the 2011 Japan earthquake resulted in over 15,000 deaths and approximately \$235 billion in economic losses, while the 2023 Turkey-Syria earthquake caused more than 50,000 fatalities \citep{sun2022novel}. Between 2016 and 2020, disasters affected over 602 million people globally \citep{yang2023distributionally}. Timely post-disaster emergency responses that include the delivery of essential supplies such as food, water, and medical aid are critical for saving lives \citep{world2000natural}. However, these efforts are often hindered by uncertainties in transportation infrastructure disruptions immediately following a disaster. Such uncertainties encompass the extent of road network damage, current operability, and feasibility of short-term restoration. Thus, rapid assessment of road network damage is essential to mitigate this uncertainty and enable effective humanitarian interventions.

Unmanned aerial vehicles, or drones, have emerged as a transformative tool for post-disaster damage assessment, offering distinct advantages over traditional methods \citep{enayati2023multimodal,shi2024optimal,van2025stochastic}. Their ability to capture high-resolution imagery at lower operational costs than satellites or helicopters makes them well-suited for rapid deployment in disaster zones \citep{zhang2023robust}. Equipped with advanced sensors such as high-definition cameras, drones can systematically collect detailed data on road network conditions, including damage type and severity (e.g., cracks, landslides or complete collapses), current trafficability, and feasibility of short-term repairs. For instance, during the 2013 Typhoon Haiyan response in the Philippines, drones rapidly mapped affected areas and identified blocked roads, enabling relief teams to prioritize clearance efforts \citep{turk2014drones}. In Nepal’s 2015 earthquake recovery, thermal drones assisted in locating survivors and assessing hard-to-reach mountain roads, showcasing their versatility in complex terrains \citep{abuali2025innovative}. In post-disaster scenarios, traditional transportation options like cars, motorcycles, and even pedestrian access often become impassable and hazardous \citep{zhang2023robust}. Roads may be collapsed or blocked by landslides, making ground-based access extremely difficult or even life-threatening. Drones, however, can operate freely in such hazardous environments, bypassing damaged infrastructure to collect critical information \citep{enayati2023multimodal,shi2024optimal,van2025stochastic}. This agility is pivotal in disaster response, where time sensitivity means timely information can determine survival outcomes. Building on these validated applications, this study proposes deploying a fleet of drones for rapid post-disaster road damage assessment.

Post-disaster road damage assessment via drone encompasses two critical timeframes: 1) the duration required to determine the routes for road damage assessment, and 2) the time taken by each drone to traverse its designated route. The first timeframe focuses on rapidly determining drone routing paths to enable efficient drone deployment. Existing literature tends to rely on traditional methodologies, including heuristics (e.g., genetic algorithms and simulated annealing) or exact methods (e.g., branch-and-price and column generation), to solve this routing problem \citep{huang2013continuous, balcik2020robust}. However, these methods suffer from three key limitations. First, their computational time increases significantly with problem size, making them ill-suited for time-critical disaster scenarios where delays in providing solutions can be unacceptable \citep{luo2023neural, liu2024evolution}. Second, developing a universally efficient algorithm to cope with such a drone routing problem is challenging. Such methods typically require extensive domain-specific expertise to engineer problem-tailored heuristics or exact formulations, which leads to prohibitive development costs and prolonged deployment cycles. Third, the problem’s constraints evolve rapidly with changing environments \citep{kwon2020pomo, liu2024evolution}, thus necessitating frequent redesign of specialized algorithms. 

Advancements in computer vision (CV) and large language models (LLMs) have revolutionized their respective fields: expert-driven manual feature engineering has been supplanted by automated end-to-end deep learning, enabling models to learn complex patterns and generalize across diverse scenarios without heavy human intervention \citep{brown2020language, radford2021learning, luo2023neural, zhou2024mvmoe, liu2024multi, berto2024routefinder, huang2025orlm}. Drawing inspiration from this paradigm shift, this study adopts a similar philosophy by proposing a deep reinforcement learning (DRL) model. Like CV and LLM that autonomously extract meaningful representations from raw data, the DRL model eliminates the need for domain-specific algorithm design by learning optimal routing strategies directly from problem instances. This allows it to determine high-quality drone assessment routes within seconds, adapting to varying disaster scenarios and outperforming traditional algorithms by a significant margin. The second timeframe ensures timely completion of assessments from route initiation to drone return. The proposed model explicitly incorporates this timeliness constraint, optimizing both route planning and completion time within a predefined limit. To achieve the above objectives, this study presents the following contributions:
\begin{itemize}
\item We propose an attention-based encoder-decoder model (AEDM) for rapid drone route planning in post-disaster road assessment. Unlike traditional methods requiring domain-specific algorithm design, AEDM learns optimal routing strategies autonomously through deep reinforcement learning.

\item We develop a network transformation and synthetic data generation framework that converts link-based routing into node-based formulation while generating diverse training instances, addressing the dual challenges of problem complexity and dataset scarcity.

\item We propose a multi-task reinforcement learning framework with exponential moving average based reward normalization that enables simultaneous training across varying drone numbers and time constraints, eliminating the need for separate models per parameter combination.

\item We conduct extensive experiments demonstrating that AEDM achieves 20--71\% improvements over commercial solvers and 23--35\% over traditional heuristics while maintaining rapid inference (1--2 seconds), with strong generalization to unseen parameters and real-world networks.
\end{itemize}

The paper is structured as follows: Section~\ref{Related work} reviews the relevant literature. Section~\ref{Problem definition} introduces the drone-based road network assessment problem. Section~\ref{Network transformation} presents a simple yet efficient network transformation approach. Section~\ref{Attention-based encoder-decoder model} proposes AEDM. Section~\ref{Experiments} evaluates the model's performance. Finally, Section~\ref{Conclusion} provides a summary.

\section{Related work}
\label{Related work}

In prior road damage assessment practices, four primary approaches have been employed. The first approach entails field surveys, which consist of on-site investigative activities such as community mapping and transect walks \citep{maya2013rapid}. While these methods can provide detailed data, they are time-consuming, labor-intensive, and often fail to deliver timely insights during the critical early phases of disaster response \citep{tatham2009investigation}. The second approach relies on remote sensing and geographic information system-based methods. These utilize satellite imagery to rapidly identify impacted areas. However, satellite revisit intervals (ranging from days to weeks) and resolution constraints limit their capacity to support prompt responses and detect fine-grained damages \citep{bravo2019use}. The third approach makes use of traffic data and sensor networks. It leverages global positioning systems, cameras, and embedded sensors to infer real-time road conditions. Yet, the limited deployment of such sensors in remote or underdeveloped regions undermines their feasibility for large-scale disaster scenarios \citep{yang2023distributionally}. The fourth and final approach features drones. Drones emerge as a highly promising solution, owing to their high mobility, rapid deployability, and ability to capture high-resolution data in hazardous environments \citep{agatz2018optimization,zhang2023robust}.

Recent literature has explored drone applications in post-disaster assessment. For example, \cite{huang2013continuous} formulated an assessment routing problem focusing on minimizing the total response time for visiting affected communities. \cite{balcik2017site} and \cite{balcik2020robust} extended this to optimize site selection and routing under uncertain travel conditions, using traditional optimization techniques such as mixed-integer linear programming (MILP) solvers, genetic algorithms, or tabu search. However, these methods require significant computational time as problem size increases, making real-time deployment challenging. \cite{oruc2018post} combined drones and ground vehicles to optimize the assessment process but constrained drone movement to the physical road network, limiting flexibility. \cite{zhang2023robust} introduced a robust team-orienteering model solved via a branch-and-price approach, which remains computationally intensive for larger problems. Meanwhile, \cite{adsanver2025predictive} applied a multistage framework for grid damage assessment, but their heuristic-based models exhibit increased solution time with problem complexity. Table \ref{Table 1} presents a comparative overview of these methods, highlighting key differences in solution approaches, computational performance, and reliance on domain knowledge. It shows that traditional methods, whether heuristic (e.g., genetic algorithms and tabu search) or exact (e.g., branch-and-price), exhibit solution time that increases significantly with problem size even for moderate scales (e.g., 100+ nodes) and inherently require extensive domain-specific expertise for algorithm design. In contrast, our proposed DRL-based approach achieves rapid performance while handling larger networks (up to 1,000 nodes) and eliminates the need for domain knowledge, addressing the critical time sensitivity and expertise barriers of disaster response.

Given the time sensitivity of disaster response, the determination of assessment routes for each drone post-disaster must be as rapid as possible, ideally in real time. Heuristic algorithms, which aim to balance computational efficiency and solution quality, exhibit three notable limitations: First, they require substantial domain expertise to design effective rules, rendering the process time-consuming and costly \citep{liu2022flying,luo2023neural,zhou2024mvmoe,liu2024multi}. Second, they lack flexibility when faced with environmental changes or varying problem types, necessitating frequent adjustments \citep{zhang2020multi, berto2024routefinder}. Third, while performing adequately on smaller-scale problems, they incur longer solution time for larger scenarios, reducing computational efficiency and making them ill-suited for rapid applications \citep{chen2022deep, arishi2023multi}. As a complement to these existing approaches, this study specifically addresses the urgency of post-disaster assessment by proposing a DRL model that determines high-quality drone routing plans within seconds without domain knowledge. Our approach leverages advances in artificial intelligence (AI) to overcome the scalability and flexibility limitations inherent in classical optimization and heuristic methods, thereby enabling rapid and efficient damage assessment in complex post-disaster environments.

\begin{table}[!htbp]
\small
\centering
\caption{Comparison of Road Damage Assessment Methods and Solutions}
\fontfamily{ptm}\selectfont
\label{Table 1}
\scalebox{0.8}{
\begin{tabular}{lllllccc}
\toprule
\textbf{Authors} & 
\textbf{Tools} & 
\textbf{Objectives} & 
\textbf{Solution} & 
\textbf{Type} & 
\textbf{Scale} & 
\textbf{Time} & 
\textbf{Expertise} \\
\midrule
\cite{huang2013continuous} & Route teams & Minimize arrival time & Continuous approx. & Heuristic & 10--20 teams & Increases & Required \\
\cite{balcik2017site} & Route teams & Balanced coverage & Tabu search & Heuristic & 94 nodes & Increases & Required \\
\cite{balcik2020robust} & Route teams & Balanced coverage (uncertain) & Tabu search & Heuristic & 94 nodes & Increases & Required \\
\cite{oruc2018post} & Drones \& bikes & Max. link \& node value & Base route & Heuristic & 44 nodes & Increases & Required \\
\cite{glock2020mission} & Drones & Max. informativeness & ALNS$^{a}$ & Heuristic & 625 nodes & Increases & Required \\
\cite{zhang2023robust} & Drones & Max. info. within time & Branch-\&-price & Exact & 107 nodes & Increases & Required \\
\cite{yin2023robust} & Trucks \& drones & Min. costs (uncertain) & Branch-price-cut & Exact & 45 nodes & Increases & Required \\
\cite{morandi2024orienteering} & Truck \& drones & Max. prize within time & Branch-\&-cut & Exact & 50 nodes & Increases & Required\\
\cite{adsanver2025predictive} & Drones & Max. priority scores & VND$^{b}$ & Heuristic & 148 nodes & Increases & Required \\
\midrule
\textbf{This study} & \textbf{Drones} & \textbf{Max. info. within time} & \textbf{DRL (AEDM)} & \textbf{Learning} & \textbf{1,000 nodes} & \textbf{1--2s} & \textbf{Not req.} \\
\bottomrule
\end{tabular}}
\begin{flushleft}
\footnotesize
\textit{Note:} $^{a}$ALNS = Adaptive Large Neighborhood Search; $^{b}$VND = Variable Neighborhood Descent. ``Increases'' indicates computation time increases significantly with problem size. Max. = Maximize; Min. = Minimize.
\end{flushleft}
\end{table}

\section{Problem definition}
\label{Problem definition} 

In the immediate aftermath of a disaster, the highly uncertain extent of road network damage underscores the critical need to rapidly assess network usability before executing emergency supply deployments and evacuation efforts. This problem is formally framed as follows. Given a road network modeled as a graph $G = (N, A)$ (where $N$ denotes the set of road nodes, encompassing intersections in urban settings or intersections/villages in rural contexts, and $A$ represents the links between these nodes) and a fleet of drones, the objective is to determine the assessment routes for each drone. These routes aim to maximize the total value of damage-related information collected across the network links. Drones, outfitted with high-definition cameras, are deployed to evaluate the damage status of each link. Due to a drone’s operating altitude of 120 meters and its visual range (forming a circular area with a radius of approximately 75–80 meters), simultaneously inspecting multiple roads is impractical even within urban environments \citep{zhang2023robust}. Consequently, a bidirectional road link can only be assessed if a drone flies directly along it.

\begin{figure}
\centering
\includegraphics[width=0.5\textwidth]{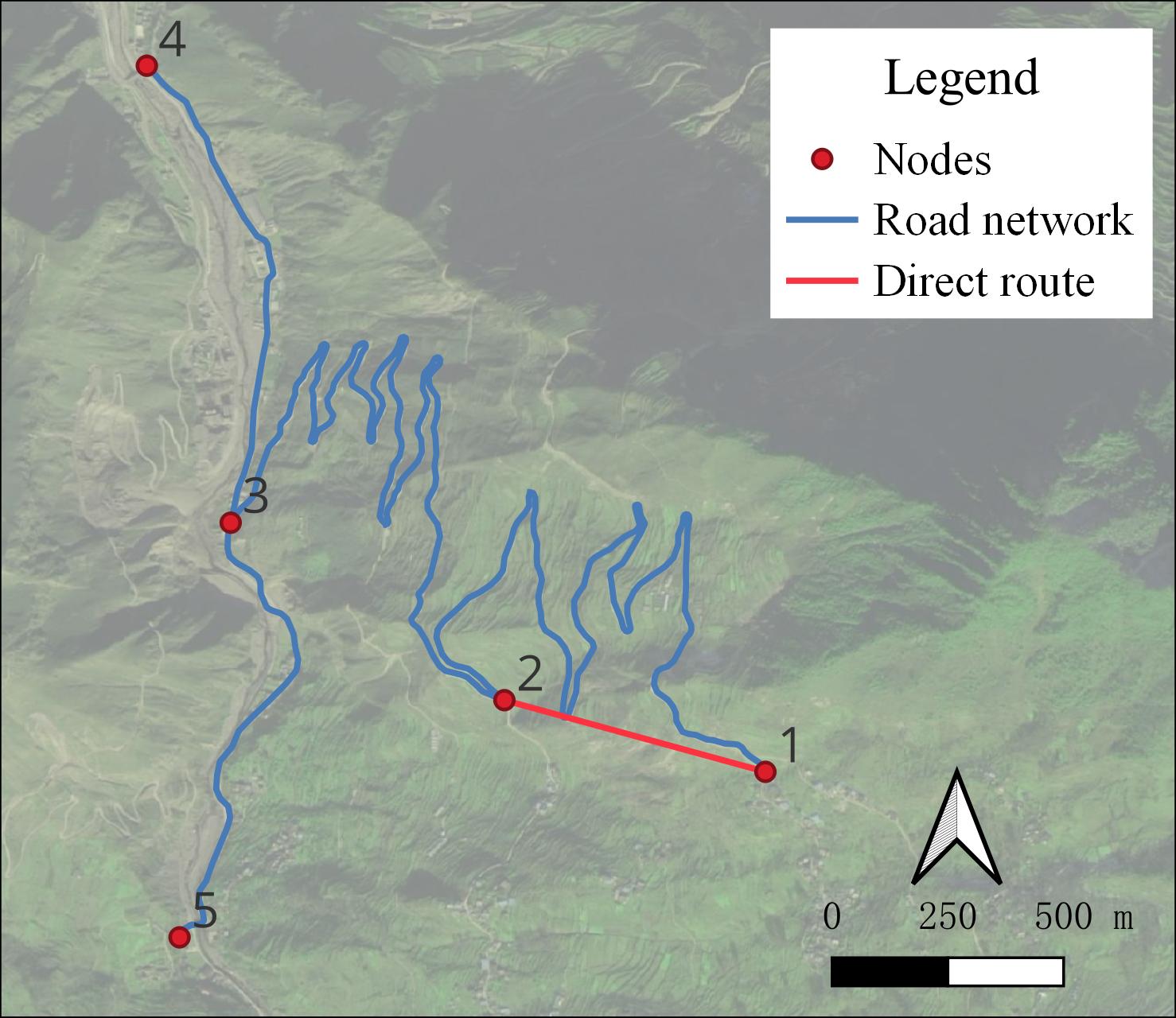}
\caption{Dual Network with Road Links for Assessment and Direct Routes for Transit}
\label{fig 1}
\end{figure}

To expedite drone movement and accelerate the assessment process, drones are permitted to travel directly between nodes, bypassing the physical road network links. As illustrated in \autoref{fig 1}, consider a road network $G = (N, A)$ with $|N| = 5$ and $|A| = 4$: a drone stationed at node 1 may either traverse a specific road link (e.g., link $(1,2)$, denoted by the blue line) to assess its condition, or opt for a more direct path (e.g., the straight-line route from node 1 to node 2, shown as the red line) to reach node 2 swiftly. In the latter scenario, the drone would forgo evaluating link $(1,2)$ and instead proceed to assess subsequent links (e.g., $(2,3)$). This dual-network framework comprises two distinct structures: the original road network $G$, where drones evaluate individual link damage, and a fully connected auxiliary network $G' = (N, A')$, in which every node is directly connected to all other nodes. The auxiliary network $G'$ enables drones to bypass physical road links, facilitating faster transit between nodes to prioritize assessments of additional critical links. Thus, alongside the original road network $G$, the introduction of $G'$ enhances traversal efficiency, balancing thorough damage assessment with time-sensitive operational needs.

Let $K$ denote the total number of drones, and $Q$ represent the battery-imposed flight time limit for each drone. For the road network $G = (N, A)$, let $(i,j) \in A$ denote an individual road link, where $i, j \in N$ are road nodes. The time required for drone $k$ to traverse link $(i,j)$ while conducting a damage assessment is denoted as $t_{ij}$, which is computed based on the drone’s flying speed and the physical length of $(i,j)$. Conversely, $t'_{ij}$ represents the transit time between nodes $i$ and $j$ in the fully connected network $G' = (N, A')$, where the drone travels directly (without assessing $(i,j)$); $t'_{ij}$ is typically shorter than $t_{ij}$ due to the direct flight path. Let $c_{ij}$ denote the damage information value obtained from assessing link $(i,j)$. Given the time sensitivity of post-disaster assessments, $p_{\max}$ is defined as the maximum allowable duration for the entire assessment process. Under this constraint, each drone must complete its assigned assessments and return to its starting depot within $p_{\max}$.

The dual-network setup outlined above introduces a key challenge: ambiguities in distinguishing between road links (for assessment) and direct routes (for transit) when both connect the same pair of nodes. This ambiguity complicates the formulation of routing problems. To address this, the following section presents a network transformation approach that converts the link-based routing problem into an equivalent node-based formulation, resolving such ambiguities and enabling efficient computation.

\section{Network transformation}  
\label{Network transformation}

\subsection{Transformation method}  

\begin{figure}
\centering
\includegraphics[width=0.85\textwidth]{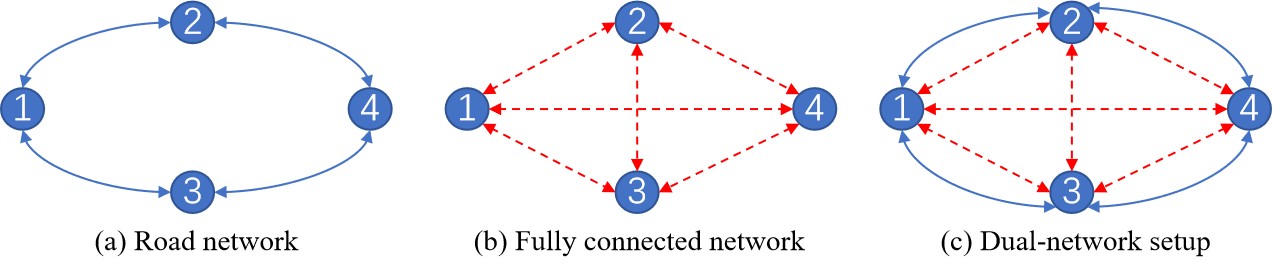}
\caption{Dual Network Ambiguity with Overlapping Links Between Nodes}
\label{fig 2}
\end{figure}

This problem involves two distinct networks: the road network $G = (N, A)$ and the fully connected node network $G' = (N, A')$. As visualized in \autoref{fig 2}, (a) depicts the original road network $G$, where nodes are connected only via physical road links (e.g., links between adjacent intersections), forming the structure where drones conduct damage assessments. (b) shows the fully connected auxiliary network $G'$, in which every node is directly linked to all other nodes, representing the direct transit paths drones can take to bypass road links. However, their integration into a dual-network configuration $G'' = (N, A + A')$ introduces ambiguities, as shown in (c): multiple links (road links vs. direct links) may connect the same pair of nodes, such as link $(1,2)$, which could represent either a road link from $G$ or a direct link from $G'$.

\begin{figure}
\centering
\includegraphics[width=0.75\textwidth]{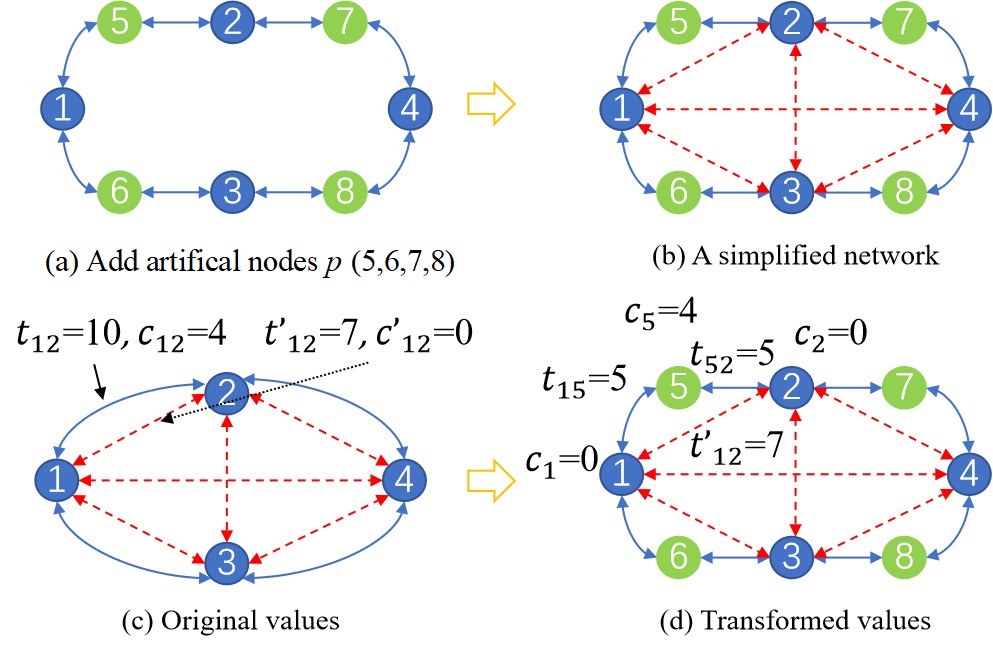}
\caption{Link-to-Node Transformation via Artificial Nodes (where 5,6,7,and 8 are artificial nodes.)}
\label{fig 3}
\end{figure}

To resolve this, existing studies have proposed transforming link-based routing networks into equivalent node-based ones. For example, \cite{pearn1988new} converted a network with $|A|$ links into an equivalent structure with $3|A| + 1$ nodes, while \cite{baldacci2006exact} and \cite{longo2006solving} developed methods to reduce this to $2|A| + 1$ nodes. However, these approaches often generate redundant links, increasing computational complexity.  
For the drone path planning problem, we adopt a traffic assignment method using intersection construction \citep{blubook} to reformulate the problem as a node-based routing task. As shown in \autoref{fig 3} (a), each road link $(i,j) \in A$ is split by an artificial node $p \in \mathcal{P}$, creating two new links: $(i,p)$ and $(p,j)$. This transforms the dual network $G'' = (N, A + A')$ into a reduced network $\bar{G} = (\bar{N}, \bar{A}) = (N \cup \mathcal{P}, 2A \cup A')$, where $|\mathcal{P}| = |A|$ (see \autoref{fig 3} (b)). In $\bar{G}$, the ambiguity of multiple links between nodes is resolved: traversing $(i,p)$ and $(p,j)$ corresponds to assessing the original link $(i,j)$, while direct links in $A'$ remain unchanged. By setting the information value of $p$ as $c_p = c_{ij}$ and splitting the traversal time as $t_{ip} = t_{pj} = t_{ij}/2$, the link-based routing problem is converted into a node-based one (examples in \autoref{fig 3} (c) and \autoref{fig 3} (d)). Notably, since $|A| \ll |A'| + |N|$ \citep{blubook}, this transformation adds only $|A|$ nodes and $|A|$ links, resulting in lower complexity compared to existing methods \citep{pearn1988new, baldacci2006exact, longo2006solving}.

\subsection{Coordinate generation}  

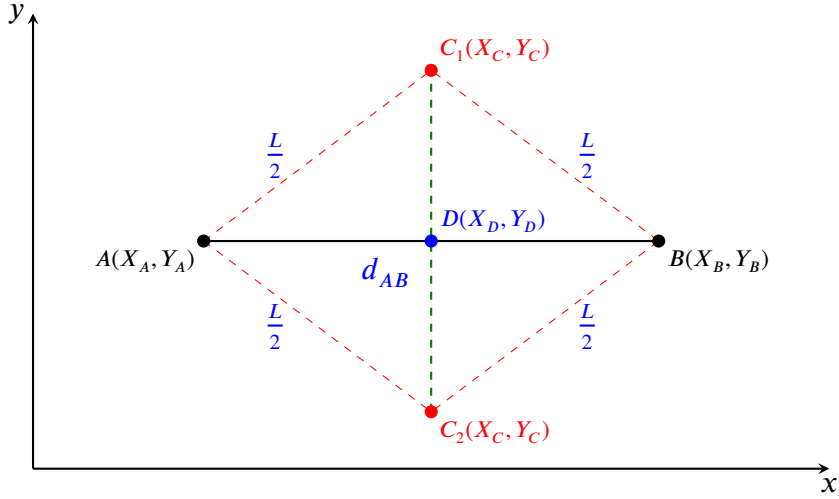
\begin{figure}
\centering
\begin{tikzpicture}[scale=1.5, >=stealth]

\draw[->, thick] (-1.5,-2) -- (5.5,-2) node[below, font=\large] {$x$}; 
\draw[->, thick] (-1.5,-2) -- (-1.5,2) node[left, font=\large] {$y$};  

\coordinate (A) at (0,0);
\coordinate (B) at (4,0); 
\coordinate (D) at (2,0); 

\pgfmathsetmacro{\dAB}{4} 
\pgfmathsetmacro{\L}{5}
\pgfmathsetmacro{\h}{sqrt(\L^2 - \dAB^2)/2}
\coordinate (C1) at (2,\h);  
\coordinate (C2) at (2,-\h); 

\draw[thick, black] (A) -- (B);

\draw[green!50!black, dashed, thick] (2,-1.5) -- (2,1.5);

\draw[dashed, red] (A) -- (C1) -- (B);
\draw[dashed, red] (A) -- (C2) -- (B);

\filldraw (A) circle (1.5pt) node[below left]{$A(X_A,Y_A)$};
\filldraw (B) circle (1.5pt) node[below right]{$B(X_B,Y_B)$};
\filldraw[blue] (D) circle (1.5pt) node[above right]{$D(X_D,Y_D)$};
\filldraw[red] (C1) circle (1.5pt) node[above right]{$C_1(X_C,Y_C)$};
\filldraw[red] (C2) circle (1.5pt) node[below right]{$C_2(X_C,Y_C)$};

\node[blue, font=\large] at ($(A)!0.4!(B)$) [below=1mm] {$d_{AB}$};
\node[blue, font=\large] at ($(A)!0.5!(C1)$) [left=3mm] {$\frac{L}{2}$};
\node[blue, font=\large] at ($(C1)!0.5!(B)$) [right=3mm] {$\frac{L}{2}$};
\node[blue, font=\large] at ($(A)!0.5!(C2)$) [left=3mm] {$\frac{L}{2}$};
\node[blue, font=\large] at ($(C2)!0.5!(B)$) [right=3mm] {$\frac{L}{2}$};
\end{tikzpicture}
\caption{Artificial Node Coordinate Calculation Using Perpendicular Bisector Method}
\label{fig 4}
\end{figure}

To systematically solve the problem, we address the coordinate generation for artificial nodes. In a given road network, geographical coordinates of original nodes are predefined (as shown in \autoref{fig 3} (c)), along with the corresponding road lengths, but coordinates of artificial nodes (e.g., nodes 5, 6, 7, 8 in \autoref{fig 3} (d)) are not. We propose a method to determine their coordinates as follows: Consider two nodes $A(X_A, Y_A)$ and $B(X_B, Y_B)$ in a 2D plane, with link $(A, B)$ of length $L$. We aim to find coordinates of point $C(X_C, Y_C)$ such that $AC = BC = L/2$ and $C$ lies on the perpendicular bisector of $AB$ (see \autoref{fig 4}). Compute the midpoint $D$ of $AB$: $X_D = (X_A + X_B)/2$, $Y_D = (Y_A + Y_B)/2$. The Euclidean distance between $A$ and $B$ is $d_{AB} = \sqrt{(X_B - X_A)^2 + (Y_B - Y_A)^2}$. Since $C$ lies on the perpendicular bisector of $AB$, the direction vector of this bisector is $(-(Y_B - Y_A), X_B - X_A)$, and its unit normal vector is $(-(Y_B - Y_A)/d_{AB}, (X_B - X_A)/d_{AB})$. The coordinates of $C$ are derived as (detailed derivation in Appendix~\ref{Derivation of Point $C$ Coordinates}):  

\begin{equation}
\label{1}
X_C = \frac{X_A+X_B}{2} \mp \frac{\sqrt{L^2-d_{AB}^2}}{2}\frac{Y_B-Y_A}{d_{AB}}
\end{equation}

\begin{equation}
\label{2}
Y_C = \frac{Y_A+Y_B}{2} \pm \frac{\sqrt{L^2-d_{AB}^2}}{2}\frac{X_B-X_A}{d_{AB}}
\end{equation}

This yields two symmetric solutions for $C$, which lie on either side of the perpendicular bisector of $AB$. One solution is uniformly selected as the coordinate of $C$. Using this method, distances between nodes are computed as 2D Euclidean distances, simplifying the generation of subsequent training instances.

\begin{figure}
\centering
\includegraphics[width=0.6\textwidth]{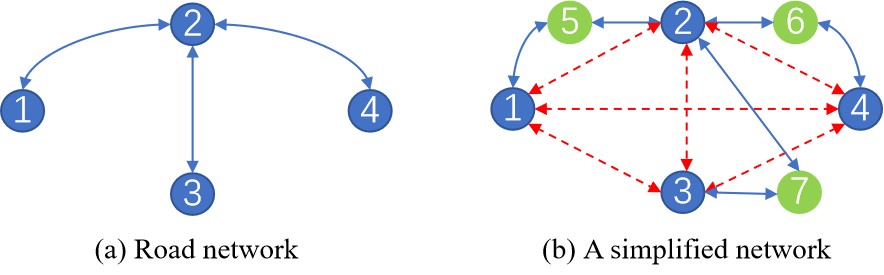}
\caption{Drone Path with Allowable Loops}
\label{fig 5}
\end{figure}

A key insight is that, via the proposed transformation (converting link-based routing to node-based routing), this drone routing problem resembles the orienteering problem (OP), where nodes have prizes, and the goal is to maximize total collected prizes while keeping route length within a maximum limit \citep{gunawan2016orienteering, kool2018attention, zhang2023robust}. However, critical differences exist as listed below, precluding direct application of existing OP algorithms. First, original nodes in $N$ carry no information value, while artificial nodes in $\mathcal{P}$ (representing road links) do. Second, in the transformed network $\bar{G}$, all nodes in $N$ are mutually connected, but artificial nodes in $\mathcal{P}$ are isolated from each other; each connects only to two specific nodes in $N$. Third, and most notably, artificial nodes in $\mathcal{P}$ are exclusive: each can be visited by only one drone (as a road link's information value is collected once), whereas nodes in $N$ (valueless) can be revisited. For example, in \autoref{fig 3} (b) with four drones (each limited to one link by time constraints), if nodes 2 and 3 were non-revisitable, only two links' information could be collected, which is suboptimal compared to four with revisits allowed. Additionally, revisits enable loops in drone paths: in \autoref{fig 5}, a single drone's optimal route (1,5,2,3,7,2,6,4) includes a loop (2,3,7,2), which would be impossible under non-revisitable constraints. This contrasts with OP's Miller-Tucker-Zemlin (MTZ) constraints \citep{miller1960integer}, which prohibit cycles to ensure path continuity. These structural differences mean existing OP algorithms (exact or heuristic) cannot be directly applied. This highlights a common limitation of traditional algorithmic solutions: even minor modifications to the problem structure often require substantial redesign, which can be time-consuming and heavily reliant on expert domain knowledge. Moreover, in disaster-response scenarios where rapid decision-making is critical, there is an urgent need for methods that can deliver high-quality solutions as swiftly as possible, ideally in real time. In the following sections, we present a novel approach designed to tackle these challenges.

\section{Attention-based encoder-decoder model}
\label{Attention-based encoder-decoder model}

To address the rapid solution requirement and eliminate the need for domain knowledge in algorithm design, we propose a learning-based framework, where a deep neural network is employed to map problem instance input data (denoted as vector $\mathbf{x}$) to a valid solution (denoted as vector $\mathbf{y}$). Given the absence of large-scale labeled training data $(\mathbf{x},\mathbf{y})$ for this novel problem, which is a common challenge for many emerging combinatorial optimization problems, we adopt reinforcement learning (RL) to train the deep neural network, enabling rapid inference without domain-specific expertise.

\begin{figure}
\centering
\includegraphics[width=0.98\textwidth]{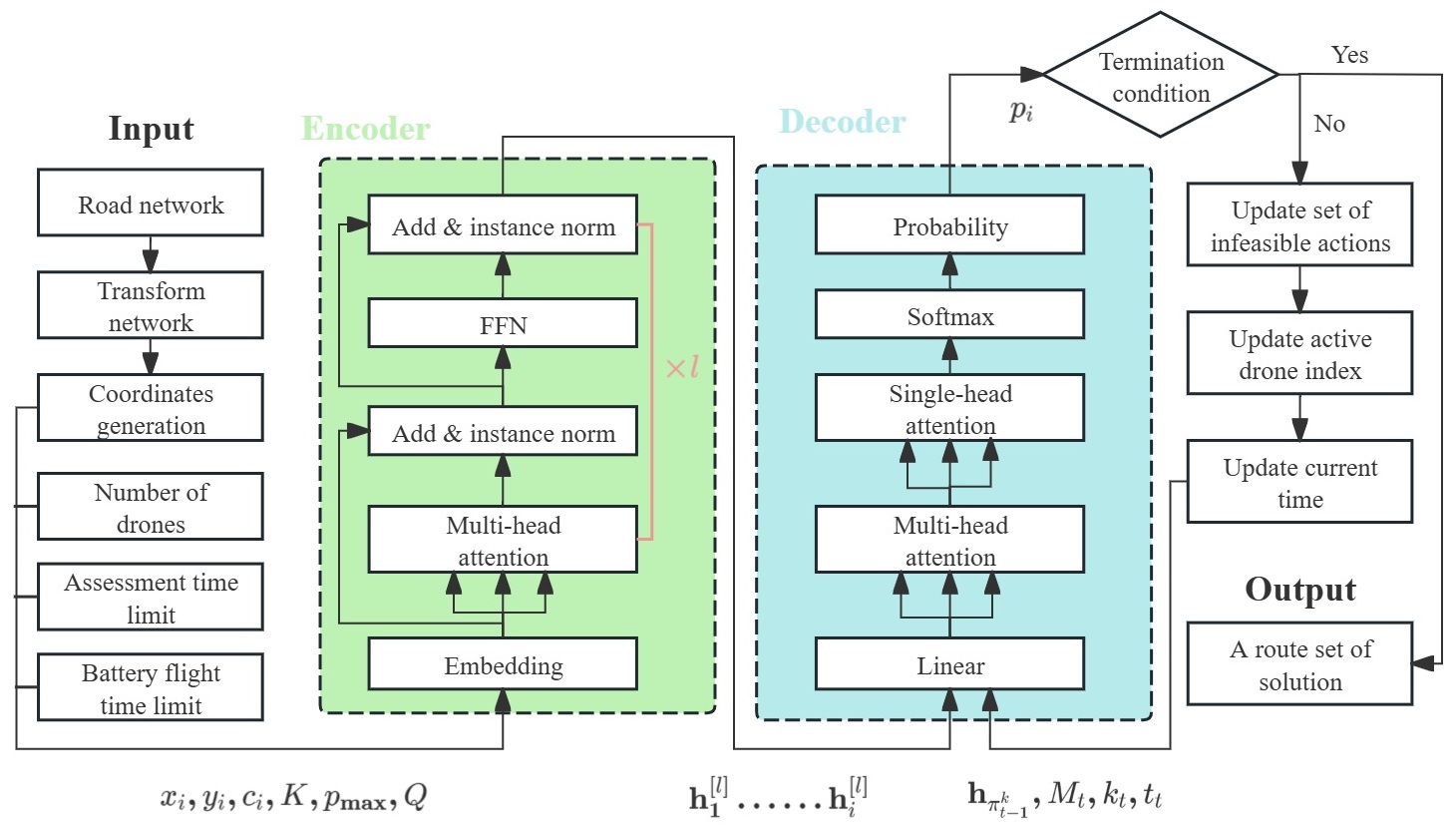}
\caption{Architecture of AEDM. The encoder processes road network coordinates and problem parameters through transformer layers to generate node embeddings. The decoder sequentially constructs drone routes using MHA and masking mechanisms to ensure feasible solutions.}
\label{fig 6}
\end{figure}

Formally, the problem is modeled as follows: for a given road network instance $s$, the network transformation method results in a road network with $\left|\bar{N}\right|$ nodes. The task involves dispatching $K$ drones to visit nodes in the transformed network $\bar{G}$, with the primary objective of maximizing the total information value collected, subject to the constraints that each drone has the battery flight time limit $Q$ and the maximum allowable time for assessment $p_{\max}$. The solution is represented by a route set $\pi = (\pi^1, \pi^2, \cdots, \pi^k)$, where $\pi^k$ denotes the sequential path of nodes visited by the $k$-th drone. We propose AEDM that defines a stochastic policy $p_{\theta}(\pi|s)$ to determine a route set solution $\pi$ for a given road network instance $s$. This policy is factorized and parameterized by $\theta$ as follows:
\begin{equation}
\label{11-1}
p_{\theta}(\pi|s)=\prod_{k = 1}^{K}p_{\theta}(\pi^k|s, \pi^{1:k-1})
\end{equation}
where $\pi^{1:k-1}$ denotes the set of paths constructed by the first $k-1$ drones, affecting the current state when generating the $k$-th drone’s path. Expanding further, for each path $\pi^k = (\pi_1^k, \pi_2^k, \cdots, \pi_{t}^k)$ where each $\pi_{t}^k$ represents each node in the path $\pi^k$, its probability is expressed as:
\begin{equation}
\label{11}
p_{\theta}(\pi^k|s, \pi^{1:k-1})=\prod_{t = 1}^{n}p_{\theta}(\pi_t^k|s, \pi^{1:k-1}, \pi_{1:t-1}^k)
\end{equation}
where $n$ is the number of nodes in the drone's path, and $\pi_{1:t-1}^k$ represents the sequence of nodes visited by the $k$-th drone before the $(t-1)$-th step. The model computes $p_{\theta}(\pi_t^k|s, \pi^{1:k-1}, \pi_{1:t-1}^k)$ by considering the current state of the drone, the status of visited nodes (from $\pi^{1:k-1}$ and $\pi_{1:t-1}^k$), and the characteristics of the current node. The overall architecture of the model is illustrated in \autoref{fig 6}, with detailed descriptions of its components presented subsequently.

\subsection{Encoder}  
\label{Encoder} 

Leveraging the transformer architecture \citep{waswani2017attention}, the encoder employs embedding techniques to process and represent input data features. In the initial encoding stage, an embedding layer maps raw node features into a higher-dimensional latent space. Nodes are partitioned into two distinct categories:

\begin{enumerate}[label=\alph*)]
    \item \textbf{Road network nodes} ($\bar{N}$): Comprising original nodes ($N$) and artificial nodes ($\mathcal{P}$), each node $i \in \bar{N}$ is defined by coordinates $(x_i, y_i)$ and an information value $c_i$. Specifically, $c_i = 0$ for $i \in N$ and $c_i > 0$ for $i \in \mathcal{P}$ (encoding road link assessment values, as detailed in Section~\ref{Network transformation}). Initial node embeddings are computed through linear transformation:
    \begin{equation}
        \mathbf{h}_i^{[0]} = [x_i, y_i, c_i] \mathbf{W}^{[0]} + \mathbf{b}^{[0]}, \quad \forall i \in \bar{N}
    \end{equation}
    where $[\cdot]$ denotes feature concatenation, $\mathbf{W}^{[0]} \in \mathbb{R}^{3 \times d}$ and $\mathbf{b}^{[0]} \in \mathbb{R}^d$ are learnable parameters, and $\mathbf{h}_i^{[0]} \in \mathbb{R}^d$ represents the initial node representation.
    
    \item \textbf{Depot node} ($o$): Functioning as the drones' origin and destination, its features incorporate coordinates $(x_o, y_o)$ and global problem parameters ($K$, $p_{\max}$, $Q$). These scalar parameters are embedded into the depot node via:
    \begin{equation}
        \mathbf{h}_o^{[0]} = [x_o, y_o, K, p_{\max}, Q] \mathbf{W}_o^{[0]} + \mathbf{b}_o^{[0]}
    \end{equation}
    where $\mathbf{W}_o^{[0]} \in \mathbb{R}^{5 \times d}$, $\mathbf{b}_o^{[0]} \in \mathbb{R}^d$ are depot-specific learnable parameters, and $\mathbf{h}_o^{[0]} \in \mathbb{R}^d$ is the initial depot embedding.
\end{enumerate}

This heterogeneous embedding strategy differentiates node types and embeds global constraints into the depot's representation, forming a holistic input for subsequent attention-based processing. These embedding are then processed by $l$ transformer layers \citep{kool2018attention}, each consisting of two sublayers: a multi-head attention (MHA) sublayer and a node-wise fully connected feed-forward network (FFN) sublayer. Inspired by residual networks, a residual connection is employed to mitigate the issues of vanishing or exploding gradients in deep neural networks \citep{he2016identity}. Additionally, instance normalization (IN) is applied to accelerate training and enhance model performance. Mathematically, a transformer layer is defined as:
\begin{equation}
\label{18}
{\bf{\bar h}}_i = {{\text{IN}}^{\left[ l \right]}}\left( {{\bf{h}}_i^{\left[ {l-1} \right]} + {\rm{MHA}}^{\left[ l \right]} \left({\bf{h}}_1^{\left[ {l-1} \right]},{\bf{h}}_2^{\left[ {l-1} \right]}, \cdots ,{\bf{h}}_i^{\left[ {l-1} \right]}\right)} \right)
\end{equation}
\begin{equation}
\label{20}
{\bf{h}}_i^{\left[ l \right]} = {{\text{IN}}^{\left[ l \right]}}\left( {{\bf{\bar h}}_i + {\text{FFN}}^{\left[ l \right]} \left({\bf{\bar h}}_i\right)} \right)
\end{equation}

After passing through $l$ transformer layers, the resulting embedding vector ${\bf{h}}_i^{[l]} \in \mathbb{R}^d$ is obtained for each node $i$, which are subsequently used as inputs to the decoder.

\subsection{Decoder}  
\label{Decoder} 

The primary function of the decoder is to sequentially construct a solution by integrating the encoder's final output embeddings $\mathbf{h}_i^{[l]}$ with the current state representation $s_t$ at each time step $t$. A context embedding mechanism dynamically modulates the query space by combining the embedding of the current problem node with relevant state information. This context embedding $\mathbf{h}_c^{(t)}$ is derived from a linear projection of concatenated features: the node embedding from the encoder's final layer ($\mathbf{h}_i^{[l]}$) and dynamic state features including the current time $d_t$ and the current index $k_t$ of the active drone. Mathematically:
\begin{equation}
\label{22}
\mathbf{h}_c^{(t)} = 
\begin{cases} 
\left[ \mathbf{h}_{\pi_{t-1}^k}, d_t, k_t \right] \mathbf{W}_c + \mathbf{b}_c & t \geq 1 \\
\left[ \mathbf{h}_{o}, d_t, k_t \right] \mathbf{W}_c + \mathbf{b}_c & t = 0
\end{cases}
\end{equation}
where $\mathbf{h}_{\pi_{t-1}^k}$ denotes the final-layer encoder embedding of the node selected at the previous timestep $\pi_{t-1}^k$; $\mathbf{h}_{o}$ represents the depot embedding from the encoder's final layer; $d_t$ and $k_t$ capture the temporal state and active drone index respectively; and $\mathbf{W}_c \in \mathbb{R}^{(d+2) \times d}$ and $\mathbf{b}_c \in \mathbb{R}^d$ are learnable projection parameters. To enhance readability, we omit the encoder layer index, and in the decoder, all variables originating from the encoder refer to the output of the final encoder layer $[l]$. This context embedding formulation enables the decoder to maintain awareness of both the current solution state and operational constraints while generating subsequent actions. The explicit incorporation of temporal progression ($d_t$) and drone index ($k_t$) allows the model to dynamically adjust its decision policy based on the evolving solution characteristics.

Next, the MHA layer and a single-head attention (SHA) layer are applied, which are expressed as:
\begin{equation}
\label{23}
{\bf{\bar h}}_c^{(t)} = {\rm{MHA}} \left({{\bf{h}}_c^{(t)}, \bf{h}}_1,{\bf{h}}_2, \cdots ,{\bf{h}}_i, M_{t} \right)
\end{equation}
\begin{equation}
\label{24}
\mathbf{u} = u_o,u_1, \cdots,u_{i} = {\rm{SHA}} \left({{\bf{\bar h}}_c^{(t)}, \mathbf{h}}_1,{\mathbf{h}}_2, \cdots ,{\mathbf{h}}_i, M_{t} \right)
\end{equation}
where $M_{t}$ denotes the set of infeasible actions at the current time step $t$, which we will discussed later. To compute the output probability $p_{\theta}(\pi_t^k|s, \pi^{1:k-1}, \pi_{1:t-1}^k)$ as given in equation \eqref{11}, the probability of selecting node $i$ at time step $t$ is computed using the softmax function:
\begin{equation}
p_i = p_{\theta}(\pi_t^k|s, \pi^{1:k-1}, \pi_{1:t-1}^k) = \text{Softmax}(C \cdot \text{tanh}(\bf{u}))
\end{equation}
where $C$ serves as a clipping parameter for the tanh function, enhancing solution exploration in the search space \citep{bello2016neural, berto2024routefinder}. Upon obtaining the probability distribution $p_i$, the current time step $t$ is finalized, and the state is updated to $t+1$, allowing the decoder to construct solutions autoregressively. The decoder terminates when either of the following conditions is met, yielding a feasible solution:  
\begin{enumerate}[label=\alph*)]
    \item All drones have been deployed, and the last drone returns to the depot;
    \item All nodes with damage information values (i.e., nodes in set $\mathcal{P}$) have been visited, and the active drone returns to the depot.
\end{enumerate}
Given that the detailed architectures of MHA, IN, FFN, and SHA represent mature techniques in large language models, which are thoroughly discussed in existing literature such as \cite{waswani2017attention,kool2018attention,kwon2020pomo,luo2023neural,zhou2024mvmoe,liu2024multi,berto2024routefinder}, this study refrains from redundant technical elaboration.

The following discusses the set of infeasible actions $M_t$ that are masked (i.e., set $u_i = -\infty$ for $i \in \bar{N}$) at the current time step $t$:
\begin{enumerate}[label=\alph*)]
\item The next node is not connected to the current node at time step $t$ in the transformed road network. This masking ensures that the decoder's output adheres to the routing rules of the road network.
\item Nodes with damage information values (i.e., belonging to the set $\mathcal{P}$) have been visited at the current time step $t$. This masking prevents redundant collection of damage information values from links.
\item Each drone departs from the depot, collects a series of nodes, and must return to the depot within the specified assessment time limit $p_{\max}$ or battery flight time limit $Q$. This masking enforces the flight time constraints for each drone. Due to the structure of the transformed road network—where nodes with damage information values are only connected to nodes without damage information values (not to the depot), and nodes without damage information values are fully connected—two cases are distinguished:
\begin{itemize}
\item If the next node $i$ to be visited belongs to $N$ (i.e., no damage information value) and is connected to the depot, the node is masked if selecting $i$ would cause the total flight time to exceed $p_{\max}$ or $Q$, i.e., $i \in \left\{ d_t + t_{\pi_{t-1}^k i} + t_{i o} > p_{\max}\right\}$ or $i \in \left\{d_t + t_{\pi_{t-1}^k i} + t_{i o} > Q \right\}$, where $t_{\pi_{t-1}^k i}$ denotes the flight time from the previous node $\pi_{t-1}^k$ to $i$, and $t_{i o}$ the flight time from the node $i$ to depot $o$, as defined in Section~\ref{Problem definition}. 
\item If the next node $i$ to be visited belongs to $\mathcal{P}$ (i.e., with damage information value) and is not connected to the depot, an additional step is considered: the flight from $i$ to the next depot-connected node. As each node $i \in \mathcal{P}$ is connected to exactly two nodes in $N$ with equal distances (see equations \eqref{1}--\eqref{2}), the node is masked if selecting $i$ would cause the total flight time to exceed $p_{\max}$ or $Q$, i.e., $i \in \left\{ d_t + 2 \cdot t_{\pi_{t-1}^k i} + t_{i o} > p_{\max}\right\}$ or $i \in \left\{d_t + 2 \cdot t_{\pi_{t-1}^k i} + t_{i o} > Q \right\}$.
\end{itemize}
\end{enumerate}
These mask mechanisms ensure adherence to the transformed network's structural constraints, prevents redundant information collection, and maintains operational time limits, guaranteeing feasible drone routing solutions.

Finally, the update rules for the active drone index $k_{t}$ and current time $d_t$ at the next time step $t+1$ are as follows. For the active drone index $k_{t+1}$:  
\begin{equation}
k_{t+1} = 
\begin{cases} 
k_{t} + 1 & \text{if the selected node } i \text{ is the depot } o \\
k_{t} & \text{if the selected node } i \text{ is not the depot}
\end{cases}
\end{equation}  
This rule ensures that when a drone returns to the depot, the next drone in the fleet becomes active, facilitating sequential deployment of multiple drones. For the current time $d_t$:  
\begin{equation}
d_{t+1} = 
\begin{cases} 
0 & \text{if the selected node } i \text{ is the depot } o \\
d_t + t_{\pi_{t-1}^k i} & \text{if the selected node } i \text{ is not the depot}
\end{cases}
\end{equation}  
Here, $d_{t+1}$ resets to 0 when returning to the depot, reflecting the start of a mission for the next drone. When visiting a non-depot node $i$, the flight time $t_{\pi_{t-1}^k i}$ from the previous node $\pi_{t-1}^k$ to $i$ is accumulated, maintaining the temporal consistency of the drone's mission timeline. These updates enable the decoder to dynamically manage multi-drone operations and enforce time constraints in the routing solution.

\subsection{Training method}  
\label{Training method}

The stochastic policy in equation \eqref{11} is parameterized by $\theta$ (encompassing all trainable variables of AEDM). For model optimization, policy gradient methods iteratively refine $\theta$ via gradient estimation of the expected reward. While \cite{kool2018attention} proposed the REINFORCE estimator, which uses a deterministic greedy rollout of the current best policy as baseline (updated periodically via paired t-test with $\alpha=5\%$), recent work by \cite{kwon2020pomo} showed that policy optimization with multiple optima (POMO) outperforms REINFORCE. Thus, we train AEDM using POMO.

Notably, our problem diverges from the original POMO in four key aspects. First, fixing $K$ (drone number), $p_{\max}$ (assessment time limit), and $Q$ (battery constraint) reduces the problem to single-task learning, which is analogous to training models for the fixed-vehicle number vehicle routing problem in \cite{zhang2020multi}. While simplifying training, this approach necessitates separate models for each parameter combination and requires matching models to combinations during deployment, which is inefficient. Inspired by multi-task capabilities of LLM, we advance AEDM as a multi-task framework: it is trained to handle diverse $(K, p_{\max}, Q)$ combinations and, crucially, generalizes to unseen ones.

Second, two strategies exist for training such a multi-task model: (1) Directly mixing diverse parameter combinations within individual batches \citep{berto2024routefinder}; (2) Training a fixed parameter set per batch, with varying combinations across successive batches \citep{zhou2024mvmoe,liu2024multi}. For our problem, decoder output lengths vary significantly across different combinations of $K$, $p_{\max}$, and $Q$. For instance, with fixed $p_{\max}$ and $Q$, the decoder output length for $K=4$ is nearly double that for $K=2$. The latter accelerates training by ensuring near-uniform decoder output lengths across instances within each batch. Thus, we adopt the second strategy.

Third, RL aims to maximize the expected reward. However, within this multi-task learning framework where batches are trained on diverse parameter combinations, rewards vary accordingly. For instance, with other parameters fixed, rewards obtained under $K=2$ differ significantly from those under $K=3$ or $K=4$. This discrepancy arises because a larger number of drones (higher $K$) can potentially collect more rewards by visiting additional nodes. To mitigate potential biases in the learning process, we propose a normalized reward strategy that combines exponentially moving average (EMA) with Z-score normalization, denoted as $r_{\text{norm}}^{(pc)}$ (where $pc$ represents a specific parameter combination). Specifically, we first compute the exponentially smoothed mean of rewards for each parameter combination over time:  
\begin{equation}
\label{15}
\hat{\mu}_t^{(pc)} = (1-\gamma) \cdot \hat{\mu}_{t-1}^{(pc)} + \gamma \cdot \mu_t^{(pc)}
\end{equation}  
where $\mu_t^{(pc)}$ is the mean reward of the current batch for $pc$, $\hat{\mu}_{t-1}^{(pc)}$ is the smoothed mean from the previous batch, and $\gamma \in [0,1]$ (set to 0.25 following \cite{berto2024routefinder}) controls the smoothing intensity, with initialization $\hat{\mu}_1^{(pc)} = \mu_1^{(pc)}$. We then calculate the exponentially smoothed variance to capture reward dispersion:  
\begin{equation}
\label{16}
\hat{\sigma}_t^{(pc)} = (1-\gamma) \cdot \hat{\sigma}_{t-1}^{(pc)} + \gamma \cdot \sigma_t^{(pc)}
\end{equation}  
where $\sigma_t^{(pc)}$ is the variance of rewards in the current batch for $pc$, with initialization $\hat{\sigma}_1^{(pc)} = \sigma_1^{(pc)}$. The normalized reward is finally computed via Z-score transformation using the smoothed statistics:  
\begin{equation}
\label{17}
r_{\text{norm}}^{(pc)} = \frac{r^{(pc)} - \hat{\mu}_t^{(pc)}}{\sqrt{\hat{\sigma}_t^{(pc)}} + \epsilon}
\end{equation}  
where $\epsilon \approx 10^{-8}$ avoids division by zero. This approach simultaneously tracks changes in both mean and variance across batches, ensuring stable multi-task convergence by anchoring rewards to their historical trajectories within each parameter regime. As validated by the ablation study in Appendix~\ref{Ablation study}, this normalization strategy outperforms alternatives such as batch-wise Z-score normalization and no normalization, minimizing relative gaps and enhancing model stability across diverse parameter combinations.

Fourth, for POMO implementation \citep{kwon2020pomo}, we sample a solution set $\Pi$, where each $\pi \in \Pi$ is determined by AEDM. Given the drone routing problem's nature, where all drones depart from a shared depot, we enhance sample diversity by sequentially selecting $N$ nodes from the transformed road network as a secondary visit set. Due to the network's structure, nodes with non-zero damage information values are disconnected from the depot, while those with zero values are connected. Thus, $N$ nodes are sequentially chosen to form the secondary visit set for drones.

Combining these four modifications, parameter optimization is conducted using POMO with a shared baseline, formulated as follows. For notational simplicity, let $R(\pi|s) = r_{\text{norm}}^{(pc)}$ denote the normalized reward function:

\begin{equation}
\label{equation function}
{\nabla}_{\theta} J(\theta) \approx \frac{1}{|\Pi|} \sum_{\pi \in \Pi} (R(\pi|s) - b(s)) {\nabla}_{\theta} \log{p}_{\theta}(\pi|s)
\end{equation}
Here, $s$ denotes a specific problem instance; $R(\pi|s)$ represents the reward yielded by solution $\pi$, corresponding to the total collected damage information value. The shared baseline $b(s)$ is defined as $b(s) = \frac{1}{|\Pi|} \sum_{\pi \in \Pi} R(\pi|s)$. Finally, $p_{\theta}(\pi|s)$ denotes the aggregated selection probabilities of nodes at each decoder step. A summary of the AEDM training procedure is provided in Algorithm \ref{algorithm1}. Additionally, the model's learning mechanism, which transitions from memorizing input parameters to reasoning about problem structures, is further validated by the t-distributed stochastic neighbor embedding (t-SNE) interpretability analysis in Appendix~\ref{Method interpretability}, which illustrates the evolutionary trajectory of features across encoder layers and confirms the effectiveness of the proposed training framework.

\begin{algorithm}
\caption{POMO with Four Modifications for AEDM Training}
\label{algorithm1}
\begin{algorithmic}[1]
\STATE \textbf{Input:} Total epochs $E$, iterations per epoch $T$, batch size $B$, parameter combinations $\mathcal{C} = \{(K, p_{\max}, Q)\}$
\STATE \textbf{Initialize:} Policy network parameters $\theta$
\FOR{epoch = 1 to $E$}
    \FOR{step = 1 to $T$}
        \STATE Sequentially  select a parameter combination $pc = (K, p_{\max}, Q)$ from $\mathcal{C}$
        \STATE Sample a batch of $B$ problem instances $s$ with fixed $pc$
        \STATE For each $s$: sequentially  select secondary starting nodes from $N$; generate rollout sequences $\Pi$ via the $\theta$-parameterized AEDM
        \STATE Normalize rewards using EMA with Z-score normalization
        \STATE Compute shared baseline: $b(s) = \frac{1}{|\Pi|} \sum_{\pi \in \Pi} R(\pi|s)$
        \STATE Calculate policy gradient using \eqref{equation function}
        \STATE Update $\theta$ via Adam optimizer: $\theta \leftarrow \text{Adam}(\theta, \nabla_{\theta} J(\theta))$
    \ENDFOR
\ENDFOR
\STATE \textbf{Output:} Optimized AEDM parameters $\theta$
\end{algorithmic}
\end{algorithm}

\subsection{Instance generation}

\subsubsection{Training instance generation}
\label{Training instance generation}

In contrast to prior DRL studies focused on combinatorial optimization problems such as vehicle routing problem (VRP) and traveling salesman problem (TSP) \citep{kool2018attention,kwon2020pomo,luo2023neural,liu2024multi,berto2024routefinder}, which generate training instances via random node placement in a coordinate system, our drone routing problem requires road network-based instances. Given the scarcity of public road network datasets (fewer than 20, e.g., \url{https://github.com/bstabler/TransportationNetworks}), which is insufficient for effective DRL training, we propose a synthetic road network generation method.

\begin{figure}
\centering
\includegraphics[width=0.98\textwidth]{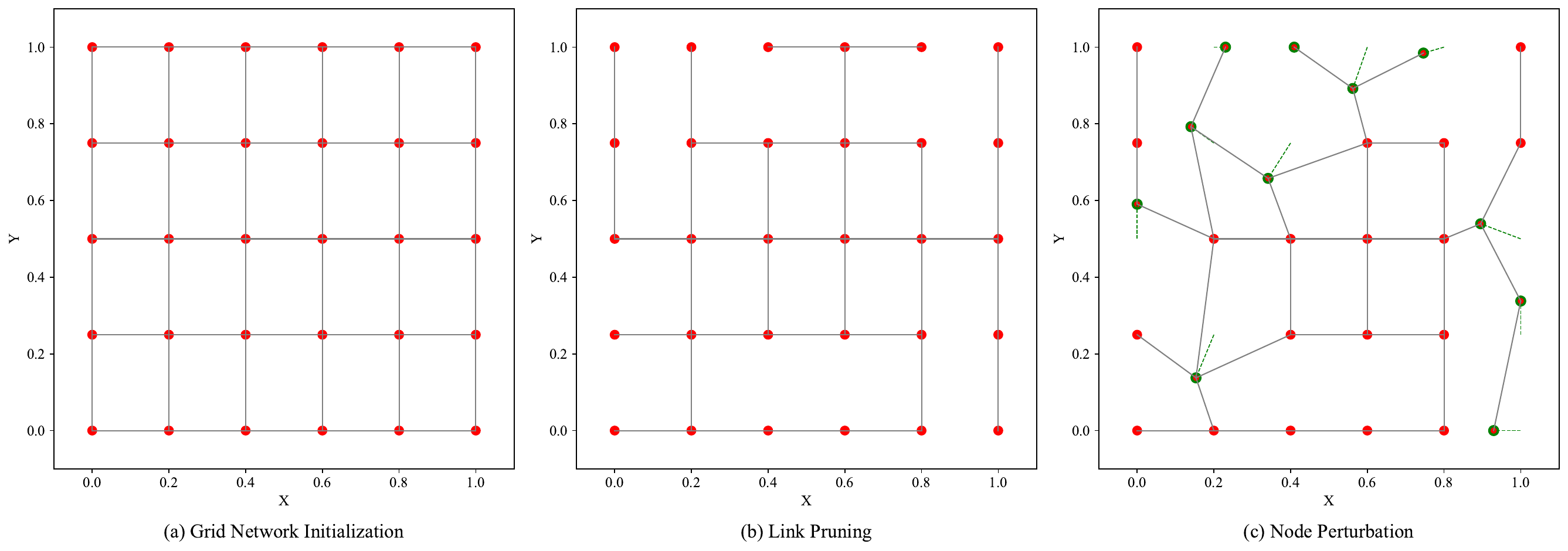}
\caption{The three-stage synthetic road network generation for drone routing training. (a) Initializes a $N=30$ node grid in $[0,1]^2$. (b) Prunes links   (prioritizing boundaries) to retain $A=36$ links while ensuring connectivity, mimicking real-world sparsity. (c) Applies bounded random node perturbations within $[0,1]^2$ to break regularity and introduce topological diversity. This pipeline addresses dataset scarcity, enabling realistic training environments for DRL-based drone routing optimization.}
\label{fig 7}
\end{figure}

Upon the occurrence of a disaster, the real-world road network $G = (N, A)$ is known, including the latitude and longitude of each node $N$, and the length of each link $A$. We can convert latitude and longitude coordinates to planar projection coordinates. These coordinates are subsequently subjected to center translation and equidistant regularization to achieve uniform normalization within the range $[0, 1]^2$, thereby preserving the proportional relationships of original distances post-normalization. Training instances are generated as follows:

\begin{enumerate}[label=\alph*)]
    \item \textbf{Grid Network Initialization}: A uniform square grid of $N$ nodes is created over $[0,1]^2$ (see \autoref{fig 7} (a)).
    \item \textbf{Link Pruning}: Randomly remove link subsets (prioritizing boundary nodes) while maintaining connectivity, resulting in $A$ links (see \autoref{fig 7} (b)).
    \item \textbf{Node Perturbation}: Introduce bounded random perturbations to node coordinates to avoid regularity, ensuring no overlaps and confinement to $[0,1]^2$ (see \autoref{fig 7} (c)).
    \item \textbf{Attribute Assignment}: Compute Euclidean lengths for remaining edges, scaled by $[1,2]$ to simulate real-world road distances. Using the transformation in Section~\ref{Network transformation}, convert to a node-based network with artificial node coordinates. Randomly sample damage information values $c_p$, serving as the reward signal during training.
\end{enumerate}

This method generates large-scale, diverse, and realistic road network instances, addressing dataset scarcity and enabling effective DRL training for disaster-response drone routing.

\subsubsection{Testing instance generation}

Given the transformed road network structure, we select $N$ nodes as the secondary visit set for drones. A critical limitation of POMO's multi-greedy inference for our problem is that $N$ (the number of greedy rollouts) cannot scale arbitrarily. Inspired by data augmentation in computer vision (e.g., cropping and flipping) and natural language processing (e.g., back-translation), \cite{kwon2020pomo} proposed generating new instances via coordinate transformations.

For our problem, we apply a similar strategy. We perform 8-fold augmentations on the node coordinates $(x_i, y_i)$ in the transformed network. These augmentations include flipping along the x-axis, y-axis, and their combinations, generating eight distinct yet equivalent problem instances. Specifically, the transformations are: keeping the original coordinates $[x, y]$; flipping the x-coordinate to get $[1-x, y]$; flipping the y-coordinate to get $[x, 1-y]$; flipping both coordinates to get $[1-x, 1-y]$; swapping coordinates to get $[y, x]$; swapping and flipping the y-coordinate to get $[1-y, x]$; swapping and flipping the x-coordinate to get $[y, 1-x]$; and swapping and flipping both coordinates to get $[1-y, 1-x]$.  

This method preserves the solution space, ensuring interchangeability between original and augmented instances. While previous works \citep{kwon2020pomo, luo2023neural, berto2024routefinder} explored additional transformations (e.g., rotations), we prioritize simplicity and omit these for this study. Future work may investigate their potential to further enhance model performance.

\section{Experiments}
\label{Experiments}

\subsection{Experiment setting}
\label{Experiment setting}

\textbf{Model Parameters.} AEDM is trained with an embedding size $d=128$. Its encoder consists of $l=6$ layers with 8-head MHA and FFN hidden size 512. These parameters choices are informed by successful applications for learning-based methods in solving traditional TSP and VRP \citep{kool2018attention,kwon2020pomo,luo2023neural,zhou2024mvmoe,liu2024multi,berto2024routefinder}. This configuration yields about 1.3 million trainable parameters. Training spans 200 epochs (batch size = 64) on 10,000 on-the-fly generated instances. The Adam optimizer uses an initial learning rate of $10^{-4}$, $L_2$ regularization ($10^{-6}$ weight decay), and learning rate decay (×0.1 at epochs 190).

\textbf{Problem Parameters.} AEDM is trained on 100-node instances, comprising 50 nodes in set $\mathcal{P}$ (with damage information values, representing 50 assessable road links) and 50 nodes in set $N$ (without damage information, including one depot). Generalization is evaluated on larger instances (200, 400, 600, 800, and 1,000 nodes), each with 50\% nodes in $\mathcal{P}$ and 50\% in $N$. Damage values for $\mathcal{P}$ are uniformly sampled from $[1, 10]$ and normalized by dividing by 10. Training uses 2, 3, or 4 drones. Practically, drones have a 2-hour battery flight time limit $Q$ and a speed of 60 km/h \citep{zhang2023robust}. Given the time-critical nature of disaster response, assessment time limits $p_{\max}$ are set to 30, 45, and 60 minutes (corresponding to 30, 45, and 60 km flight distances). Since $Q$ is significantly larger than $p_{\max}$, and drone flight time being less than $p_{\max}$ inherently implies it is less than $Q$, $Q$ is not considered in subsequent experiments, with focus on $p_{\max}$. Road network coordinates are normalized to $[0,1]$, with link lengths scaled accordingly. For simplified data generation, training uses $p_{\max} = 2, 3, 4$ corresponding to 30, 45, and 60 minutes in real-world distances.

\textbf{Training time.} Training is conducted on Google Colab using an NVIDIA A100 GPU. Each epoch requires approximately 1 minute, resulting in a total training duration of around 3.33 hours for 200 epochs. The source code for AEDM is publicly available at \url{https://github.com/PJ-HTU/AEDM-for-Post-disaster-road-assessment}.

\subsection{Benchmark}
\label{Benchmark}

The proposed problem is novel, with no existing exact or heuristic algorithms tailored to it. For comparison, we developed a corresponding mathematical model (detailed in Appendix~\ref{Mathematical formulation}) and solved it using commercial solvers (e.g., Gurobi), which is a method, like AEDM, that requires no domain knowledge for algorithm design. However, formulating a fully corresponding model for this new problem is complex, demanding domain expertise and significant time investment. To provide a more comprehensive evaluation, we additionally designed a two-phase heuristic algorithm based on classical OP frameworks \citep{kobeaga2018efficient, yang2025heuragenix}, as detailed in Appendix~\ref{Traditional optimization methods}. This heuristic consists of a greedy construction phase (GC) followed by a local-search improvement phase (LS). The construction phase uses a classic profit-to-time ratio rule to generate initial solutions, while the improvement phase applies relocate, exchange, and remove–insert operators that are carefully adapted to the link-to-node transformed network structure. Unlike the proposed AEDM, however, this heuristic requires hand-crafted domain knowledge, illustrating the typical limitations of traditional algorithmic approaches in newly emerging problem settings.

Given the time sensitivity of disaster response, both solution quality and speed are critical, and we compare both metrics across all methods: AEDM, Gurobi, and the heuristic. Notably, disaster scenarios require assessing road networks across multiple locations (e.g., post-earthquake or flood), meaning problem instances to solve are not singular but multiple. Thus, AEDM, Gurobi, and the heuristic must all handle multiple instances simultaneously. AEDM leverages GPU parallelism to process up to 1,000 instances per run, which is consistent with prior applications in combinatorial optimization (e.g., TSP and VRP) \citep{kool2018attention, kwon2020pomo, luo2023neural, zhou2024mvmoe, liu2024multi, berto2024routefinder}. While Gurobi supports parallelization across CPU cores \citep{zhou2024mvmoe, berto2024routefinder}, our experiments, which are consistent with common real-world deployment constraints due to CPU setup limitations, did not utilize this parallelization for Gurobi. The heuristic is implemented in Python and executed on the same hardware platform as other methods. To ensure fairness and reliability despite these constraints, all results are averaged over 10 problem instances for AEDM, Gurobi, and the heuristic. Specifically, AEDM's total computation time across the 10 instances is reported, while Gurobi's and the heuristic's average time per instance are used, placing the traditional methods in an advantageous position, though this does not affect AEDM's performance.

\subsection{Computational performance} 
\label{Computational performance} 

\begin{table}[!]
\caption{Performance Comparison of Methods in 200-Node Network ($p_{\max}=30$ min, Varying Drones)}
\label{Table 2}
\begin{tabular}{ccccccccc}
\toprule
Method&$^{\#}$Drone&Time (s)&Value&Gap (\%)&$^{\#}$Drone&Time (s)&Value&Gap (\%)\\
\midrule
\multirow{4}{*}{Gurobi} & \multirow{8}{*}{2} & 100 & 11.42 & 27.31 & \multirow{8}{*}{3} & 100 & 12.21 & 43.21 \\
 &  & 500 & 13.11 & 16.55 &  & 500 & 15.83 & 26.37 \\
 &  & 1,000 & 14.03 & 10.69 &  & 1,000 & 18.26 & 15.07 \\
 &  & 2,000 & 14.78 & 5.92 &  & 2,000 & 20.36 & 5.30 \\
GC &  & 1 & 11.95 & 23.93 &  & 2 & 14.24 & 33.77 \\
GC+LS &  & 18 & 12.83 & 18.33 &  & 38 & 16.68 & 22.42 \\
\rowcolor{mygreen}
AEDM&&1 & 15.07 & 4.07 &  & 1 & 20.60 & 4.19 \\
\rowcolor{myblue}
$\text{AEDM} \times 8$&&1 & 15.71 & 0.00 &  & 1 & 21.50 & 0.00 \\
\midrule
\multirow{4}{*}{Gurobi} & \multirow{8}{*}{4} & 100 & 14.92 & 43.16 & \multirow{8}{*}{5} & 100 & 14.98 & 53.75 \\
 &  & 500 & 19.13 & 27.12 &  & 500 & 21.91 & 32.36 \\
 &  & 1,000 & 19.18 & 26.93 &  & 1,000 & 22.06 & 31.89 \\
 &  & 2,000 & 20.02 & 23.73 &  & 2,000 & 23.48 & 27.51 \\
GC &  & 5 & 17.83 & 32.08 &  & 10 & 20.36 & 37.14 \\
GC+LS &  & 105 & 19.48 & 25.79 &  & 280 & 21.58 & 33.38 \\
\rowcolor{mygreen}
AEDM&&1 & 25.12 & 4.30 &  & 1 & 31.28 & 3.43 \\
\rowcolor{myblue}
$\text{AEDM} \times 8$&& 1 & 26.25 & 0.00 &  & 1 & 32.39 & 0.00 \\
\bottomrule
\end{tabular}
\end{table}

\begin{table}[!]
\caption{Performance Comparison of Methods in 200-Node Network ($p_{\max}=45$ min, Varying Drones)}
\label{Table 3}
\begin{tabular}{ccccccccc}
\toprule
Method&$^{\#}$Drone&Time (s)&Value&Gap (\%)&$^{\#}$Drone&Time (s)&Value&Gap (\%)\\
\midrule
\multirow{4}{*}{Gurobi} & \multirow{8}{*}{2} & 100 & 12.90 & 46.45 & \multirow{8}{*}{3} & 100 & 20.76 & 36.04 \\
 &  & 500 & 21.36 & 11.33 &  & 500 & 25.93 & 20.12 \\
 &  & 1,000 & 21.40 & 11.17 &  & 1,000 & 25.93 & 20.12 \\
 &  & 2,000 & 21.63 & 10.21 &  & 2,000 & 26.91 & 17.10 \\
GC &  & 2 & 16.31 & 32.29 &  & 3 & 22.68 & 30.13 \\
GC+LS &  & 28 & 19.87 & 17.52 &  & 60 & 25.76 & 20.64 \\
\rowcolor{mygreen} AEDM &  & 1 & 23.34 & 3.11 &  & 1 & 31.33 & 3.48 \\
\rowcolor{myblue} $\text{AEDM} \times 8$ &  & 1 & 24.09 & 0.00 &  & 1 & 32.46 & 0.00 \\ 
\midrule
\multirow{4}{*}{Gurobi} & \multirow{8}{*}{4} & 100 & 26.41 & 33.38 & \multirow{8}{*}{5} & 100 & 22.71 & 49.42 \\
 &  & 500 & 30.81 & 22.28 &  & 500 & 33.10 & 26.28 \\
 &  & 1,000 & 30.88 & 22.10 &  & 1,000 & 33.28 & 25.88 \\
 &  & 2,000 & 30.95 & 21.92 &  & 2,000 & 33.45 & 25.50 \\
GC &  & 7 & 28.02 & 29.31 &  & 14 & 30.94 & 31.09 \\
GC+LS &  & 145 & 31.60 & 20.28 &  & 385 & 34.91 & 22.25 \\
\rowcolor{mygreen} AEDM &  & 1 & 38.30 & 3.38 &  & 1 & 43.29 & 3.59 \\
\rowcolor{myblue} $\text{AEDM} \times 8$ &  & 1 & 39.64 & 0.00 &  & 1 & 44.90 & 0.00 \\ 
\bottomrule
\end{tabular}
\end{table}

\begin{table}[!]
\caption{Performance Comparison of Methods in 400-Node Network ($p_{\max}=30$ min, Varying Drones)}
\label{Table 4}
\begin{tabular}{ccccccccc}
\toprule
Method&$^{\#}$Drone&Time (s)&Value&Gap (\%)&$^{\#}$Drone&Time (s)&Value&Gap (\%)\\
\midrule
\multirow{4}{*}{Gurobi} & \multirow{8}{*}{2} & 100 & -- & \multicolumn{1}{c}{100} & \multirow{8}{*}{3} & 100 & -- & \multicolumn{1}{c}{100} \\
 &  & 500 & 9.41 & 62.81 &  & 500 & 10.34 & 69.03 \\
 &  & 1,000 & 17.28 & 31.70 &  & 1,000 & 20.19 & 39.53 \\
 &  & 2,000 & 20.62 & 18.50 &  & 2,000 & 25.82 & 22.67 \\
GC &  & 2 & 15.98 & 36.84 &  & 4 & 20.71 & 37.98 \\
GC+LS &  & 38 & 18.74 & 25.93 &  & 80 & 25.09 & 24.86 \\
\rowcolor{mygreen} AEDM &  & 1 & 24.30 & 3.95 &  & 1 & 32.06 & 3.98 \\
\rowcolor{myblue} $\text{AEDM} \times 8$ &  & 1 & 25.30 & 0.00 &  & 1 & 33.39 & 0.00 \\ 
\midrule
\multirow{4}{*}{Gurobi} & \multirow{8}{*}{4} & 100 & -- & \multicolumn{1}{c}{100} & \multirow{8}{*}{5} & 100 & -- & \multicolumn{1}{c}{100} \\
 &  & 500 & 15.72 & 61.60 &  & 500 & 11.41 & 75.90 \\
 &  & 1,000 & 19.96 & 51.25 &  & 1,000 & 27.17 & 42.61 \\
 &  & 2,000 & 31.84 & 22.23 &  & 2,000 & 33.47 & 29.30 \\
GC &  & 10 & 19.12 & 53.29 &  & 20 & 26.62 & 43.77 \\
GC+LS &  & 220 & 30.08 & 26.53 &  & 520 & 35.08 & 25.90 \\
\rowcolor{mygreen} AEDM &  & 1 & 39.00 & 4.74 &  & 1 & 44.91 & 5.13 \\
\rowcolor{myblue} $\text{AEDM} \times 8$ &  & 1 & 40.94 & 0.00 &  & 1 & 47.34 & 0.00 \\ 
\bottomrule
\end{tabular}
\end{table}

\begin{table}[!]
\caption{Performance Comparison of Methods in 400-Node Network ($p_{\max}=45$ min, Varying Drones)}
\label{Table 5}
\begin{tabular}{ccccccccc}
\toprule
Method&$^{\#}$Drone&Time (s)&Value&Gap (\%)&$^{\#}$Drone&Time (s)&Value&Gap (\%)\\
\midrule
\multirow{4}{*}{Gurobi} & \multirow{8}{*}{2} & 100 & -- & \multicolumn{1}{c}{100} & \multirow{8}{*}{3} & 100 & -- & \multicolumn{1}{c}{100} \\
 &  & 500 & 26.12 & 35.65 &  & 500 & 25.93 & 48.60 \\
 &  & 1,000 & 29.47 & 27.40 &  & 1,000 & 35.06 & 30.51 \\
 &  & 2,000 & 32.96 & 18.80 &  & 2,000 & 41.07 & 18.59 \\
GC &  & 3 & 25.82 & 36.39 &  & 6 & 34.82 & 30.98 \\
GC+LS &  & 58 & 33.74 & 16.88 &  & 125 & 40.48 & 19.77 \\
\rowcolor{mygreen} AEDM &  & 1 & 38.96 & 4.02 &  & 1 & 48.42 & 4.02 \\
\rowcolor{myblue} $\text{AEDM} \times 8$ &  & 1 & 40.59 & 0.00 &  & 1 & 50.45 & 0.00 \\ 
\midrule
\multirow{4}{*}{Gurobi} & \multirow{8}{*}{4} & 100 & -- & \multicolumn{1}{c}{100} & \multirow{8}{*}{5} & 100 & -- & \multicolumn{1}{c}{100} \\
 &  & 500 & 27.41 & 55.16 &  & 500 & 25.61 & 63.06 \\
 &  & 1,000 & 41.13 & 32.72 &  & 1,000 & 40.31 & 41.86 \\
 &  & 2,000 & 46.22 & 24.39 &  & 2,000 & 48.33 & 30.29 \\
GC &  & 14 & 41.40 & 32.28 &  & 28 & 47.23 & 31.87 \\
GC+LS &  & 320 & 48.88 & 20.04 &  & 840 & 52.23 & 24.65 \\
\rowcolor{mygreen} AEDM &  & 2 & 58.28 & 4.66 &  & 2 & 64.31 & 7.24 \\
\rowcolor{myblue} $\text{AEDM} \times 8$ &  & 2 & 61.13 & 0.00 &  & 2 & 69.33 & 0.00 \\
\bottomrule
\end{tabular}
\end{table}

\begin{figure}
\centering
\includegraphics[width=0.98\textwidth]{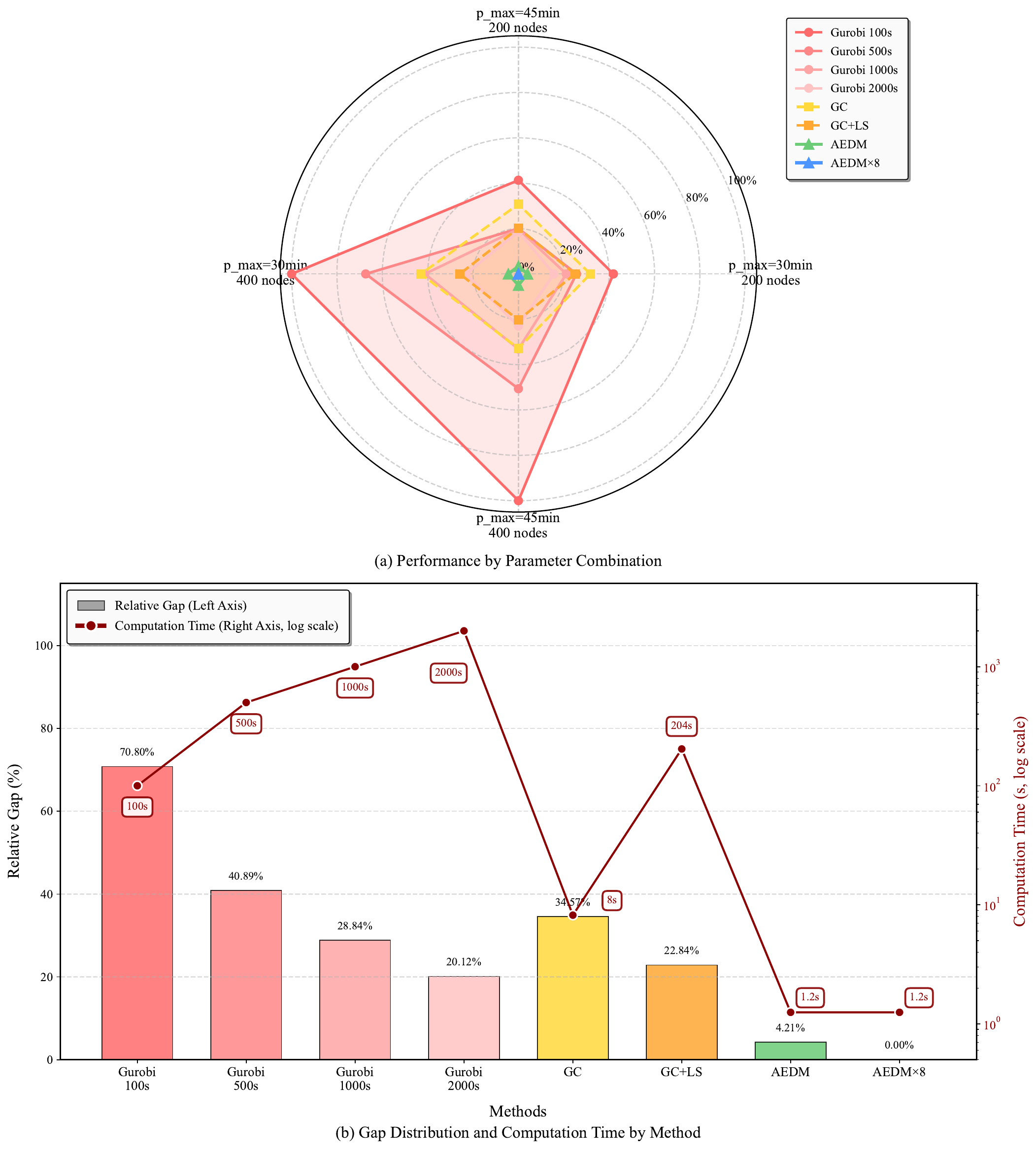}
\caption{Performance Comparison Across Methods and Parameter Combinations: (a) Relative Gap Distribution by Parameter Combination (Radar Chart); (b) Average Relative Gap and Computation Time by Method (Box Plot with Logarithmic Time Scale)}
\label{fig 8}
\end{figure}

To comprehensively evaluate the proposed AEDM, we compare it against three baseline methods: (1) Gurobi, a commercial solver that solves the mathematical formulation in Appendix~\ref{Mathematical formulation}; (2) GC (Greedy Construction), the first phase of the heuristic described in Appendix~\ref{Traditional optimization methods}; and (3) GC+LS (Greedy Construction with Local Search), the complete two-phase heuristic combining greedy construction and local search improvement. \autoref{Table 2} and \autoref{Table 3} compare results across varying numbers of drones in a fixed 200-node network, with the maximum allowable time $p_{\max}$ set to 30 and 45 minutes, respectively. \autoref{Table 4} and \autoref{Table 5} present analogous comparisons for a fixed 400-node network under the same $p_{\max}$ values, with drone numbers varied. In each table, the first column lists the method names, where AEDM~$\times$~8 denotes the proposed model with 8$\times$ instance augmentation during testing. The second column indicates the number of drones; the third column reports solution time, where Gurobi's values correspond to its maximum allowed solving time (100, 500, 1,000, and 2,000 seconds) to reflect disaster response time constraints, while GC, GC+LS, and AEDM show actual computation time per instance. The fourth column presents the objective value (total collected damage information value), and the fifth column shows the relative gap, defined as $\text{Gap} = (y - y_{\text{other}})/y \times 100\%$, where $y$ is the objective value from AEDM~$\times$~8 and $y_{\text{other}}$ from other methods.

The experimental results demonstrate clear performance advantages of AEDM across all tested scenarios. For 200-node networks, Gurobi exhibits substantial gaps even with extended computation times up to 2,000 seconds, while the heuristic GC+LS achieves improved but still inferior solutions compared to AEDM~$\times$~8. When problem scale increases to 400 nodes, Gurobi fails to return feasible solutions within 100 seconds (marked as ``--'' in \autoref{Table 4} and \autoref{Table 5}) and continues to show significant performance gaps even at 2,000 seconds. The heuristic maintains consistent gap percentages but requires computation times that escalate substantially with problem complexity.

In contrast, both AEDM and AEDM~$\times$~8 maintain consistent rapid performance within 1--2 seconds regardless of network size, drone count, or time limit configuration. The augmented version AEDM~$\times$~8 consistently delivers the best objective values through its 8-fold coordinate transformation strategy that explores diverse solution spaces with negligible computational overhead. \autoref{fig 8} provides visual confirmation through radar charts and box plots, showing that AEDM~$\times$~8 maintains minimal gaps uniformly across all parameter combinations while other methods exhibit substantially larger deficits. The box plot quantifies average performance across all scenarios, revealing that AEDM achieves only 4.21\% average gap with computation time of merely 1.2 seconds, compared to significantly larger gaps and longer times for Gurobi and GC+LS.

An important distinction between AEDM and traditional methods concerns the development effort required. As detailed in Appendix ~\ref{Mathematical formulation} and ~\ref{Traditional optimization methods}, Gurobi and the GC+LS heuristic demanded substantial domain expertise to design problem-specific operators and calibrate search parameters. In contrast, AEDM eliminates such manual algorithm engineering by learning routing policies directly from synthetic training instances. This automated learning paradigm not only achieves superior performance but also provides inherent adaptability when problem constraints evolve.

In summary, these results establish AEDM as a fundamentally superior approach for post-disaster road damage assessment, delivering three critical advantages simultaneously: (1) solution quality that surpasses both exact methods and traditional heuristics by substantial margins (20--71\% improvements over Gurobi, 23--35\% over GC+LS); (2) rapid computational efficiency with consistent 1--2 second inference regardless of problem scale; and (3) elimination of domain-specific algorithm design requirements through end-to-end learning. This combination positions AEDM as a paradigm shift from expert-driven optimization to data-driven learning for time-critical combinatorial decision-making in humanitarian logistics.

\subsection{Method robustness}
\label{Method robustness}

\begin{figure}
\centering
\includegraphics[width=0.98\textwidth]{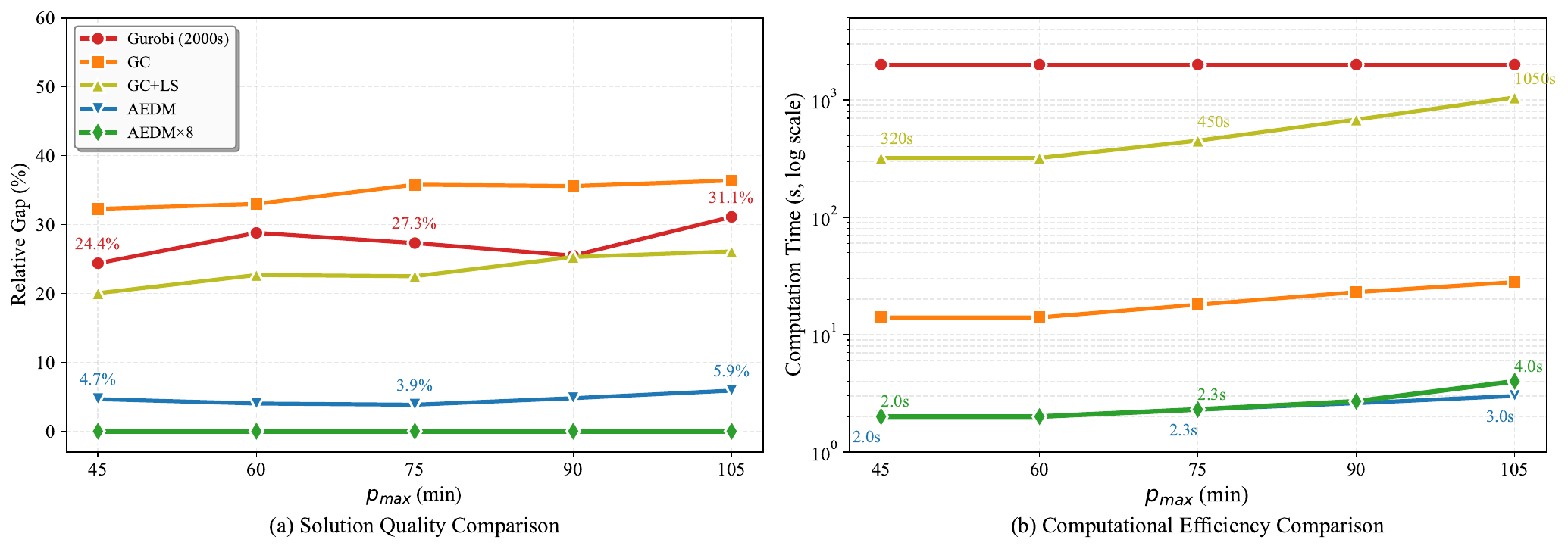}
\caption{Sensitivity Analysis with Varying Maximum Allowable Time ($p_{\max}$); Fixed Node of 400 and 4 Drones}
\label{fig 9}
\end{figure}

We further analyze the robustness of AEDM by testing it on parameter distributions that were not seen during training. \autoref{fig 9} evaluates AEDM across varying maximum allowable assessment time ($p_{\max}$), including 45, and 60 minutes that appeared during training as well as 75, 90, and 105 minutes that were not seen before. The experimental setup fixes the number of nodes at 400 and the number of drones at 4. As shown in the figure, Gurobi consistently reaches the 2,000 second time limit for all $p_{\max}$ settings. The two heuristic methods, GC and GC+LS, also exhibit rapidly increasing computational burden as $p_{\max}$ grows. GC provides fast but low quality solutions, while GC+LS improves solution quality with significantly longer runtime. In contrast, AEDM completes rapid computation for all $p_{\max}$ values and remains stable on both training seen ranges (45 to 60 minutes) and unseen ranges (75 to 105 minutes). These results confirm the temporal generalization ability of AEDM and highlight its superiority over traditional algorithms whose performance is hindered by time limits or rapidly increasing complexity.

\begin{figure}
\centering
\includegraphics[width=0.98\textwidth]{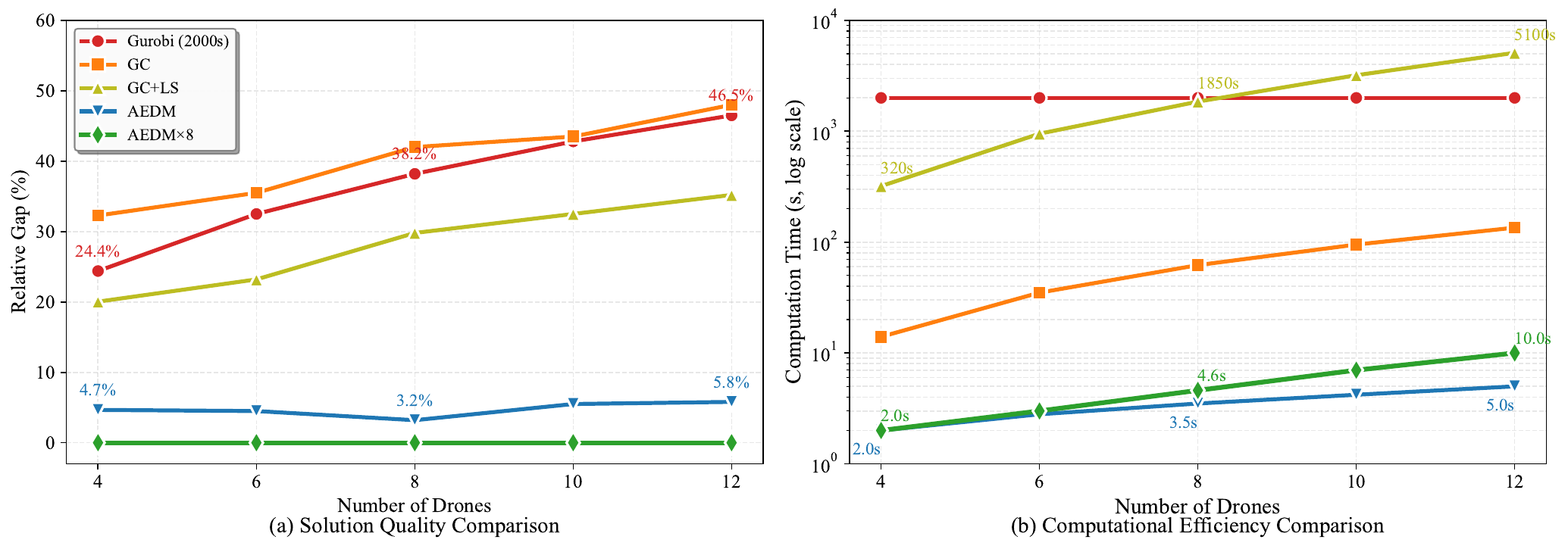}
\caption{Sensitivity Analysis with Varying Numbers of Drones; Fixed Node of 400 and $p_{\max} = 45$}
\label{fig 10}
\end{figure}

\autoref{fig 10} examines AEDM under varying drone numbers from 4 to 12, with an interval of 2. Drone number 4 is included during training while 6 to 12 are not seen before. The experimental setup fixes the number of nodes at 400 and $p_{\max} = 45$. Similar to the previous results, Gurobi reaches the 2,000 second time limit across all settings, and the heuristic methods exhibit instability as drone number increases. GC deteriorates sharply when more drones are required to coordinate, while GC+LS improves on GC but requires computation times that increase from tens to hundreds of seconds. AEDM solves all cases within seconds and maintains strong performance across both seen ranges and unseen ranges. These results demonstrate that AEDM generalizes effectively to higher coordination complexity and sustains both computational stability and solution quality when facing increased combinatorial difficulty.

\begin{figure}
\centering
\includegraphics[width=0.98\textwidth]{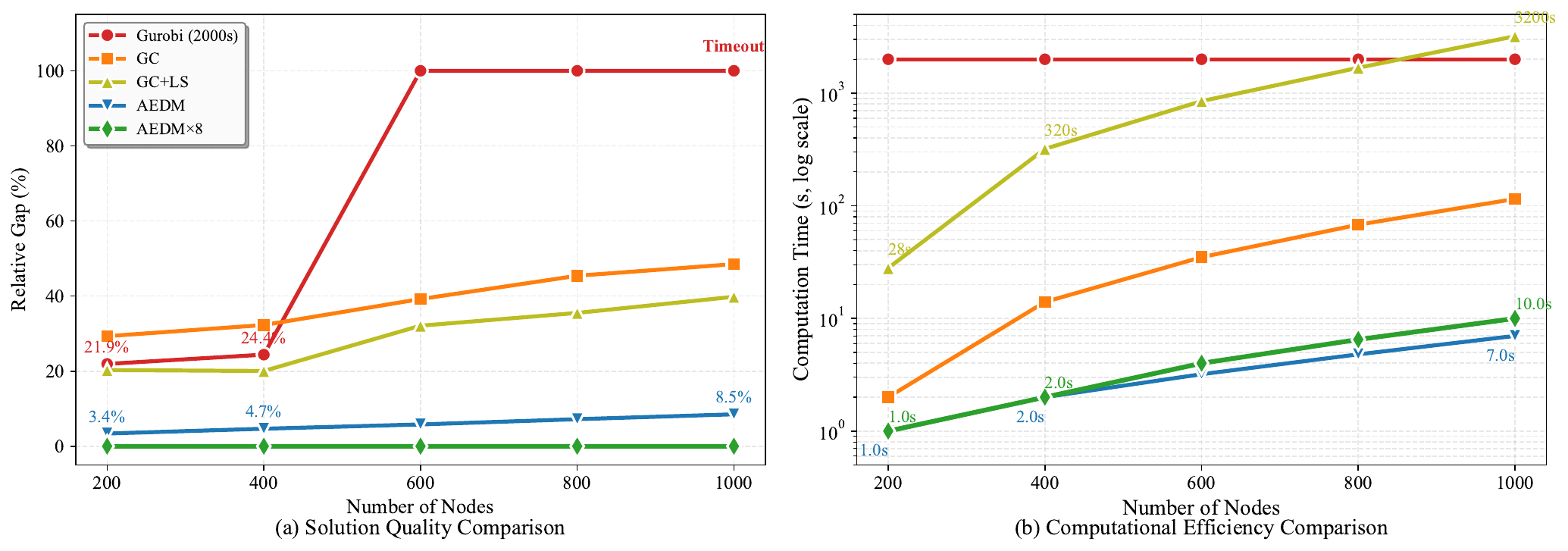}
\caption{Sensitivity Analysis with Varying Node Sizes; Fixed $p_{\max} = 45$ and 4 Drones}
\label{fig 11}
\end{figure}

\autoref{fig 11} assesses AEDM on unseen node scales ranging from 200 to 1,000 nodes, with an interval of 200. The experimental setup fixes the number of drones at 4 and $p_{\max} = 45$. For smaller scales from 200 to 400 nodes that align with training distributions, GC, GC+LS, and AEDM all produce feasible solutions, although Gurobi reaches the time limit and GC produces low quality results. GC+LS improves the solution quality over GC but with significantly longer computation time. For larger unseen scales from 600 to 1,000 nodes, Gurobi fails to return feasible solutions within the 2,000 second time limit. The heuristic methods also struggle. GC becomes unstable, and GC+LS requires extremely long computation time while still producing lower quality results. AEDM continues to generate solutions within seconds even at 1,000 nodes and produces much higher collected information values than both heuristic baselines. These results demonstrate that AEDM scales effectively to large unseen networks where traditional solvers and heuristics become infeasible.

In summary, this section evaluates AEDM across unseen parameter distributions including assessment time limits, drone numbers, and node scales. Across all settings, AEDM consistently sustains rapid computation and strong solution quality. In comparison, Gurobi and the heuristic methods GC and GC+LS exhibit significant limitations, including long computation time, unstable performance, and failure to return solutions for large scale problems. These findings confirm the strong generalization capability of AEDM and its practical advantage for time critical and large scale road network assessment tasks.

\subsection{Real-world application} 
\label{Real-world application}

\begin{figure}
\centering
\includegraphics[width=0.55\textwidth]{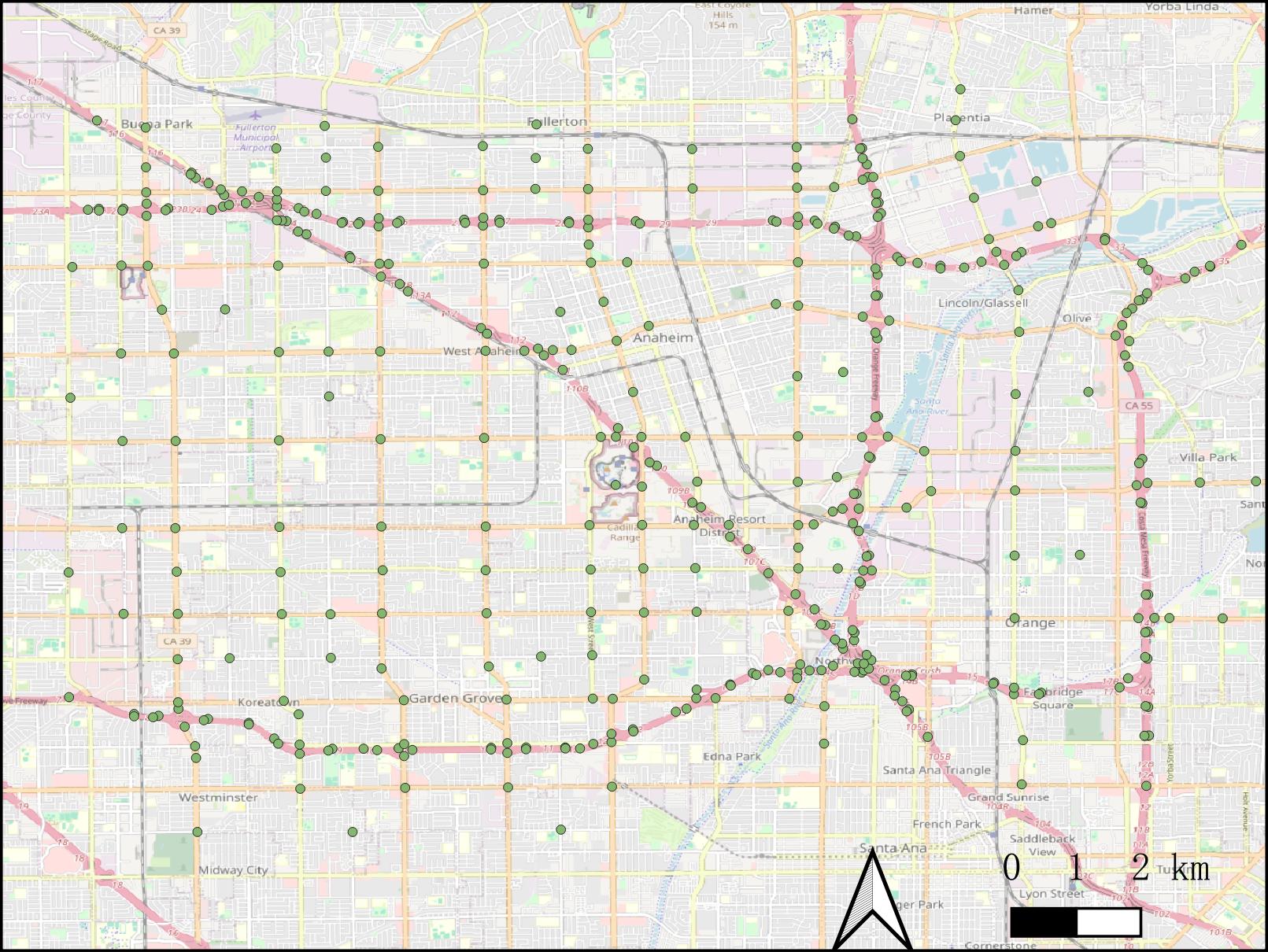}
\caption{Anaheim road network}
\label{fig 12}
\end{figure}

To validate AEDM's performance on realistic road networks, we conduct experiments using a publicly accessible transportation network from the GitHub repository (\url{https://github.com/bstabler/TransportationNetworks}), which represents a large-scale and complex scenario typical in post-disaster assessment. The Anaheim network (416 nodes, 914 links) serves as a transportation system, approximating the complexity of post-disaster road damage assessment, as shown in \autoref{fig 12}. Following the coordinate normalization procedure described in Section~\ref{Training instance generation}, we convert the geographical coordinates (latitude and longitude) of the network to planar projection coordinates, then apply center translation and equidistant regularization to achieve uniform normalization within $[0,1]^2$. 

\begin{figure}
\centering
\includegraphics[width=0.98\textwidth]{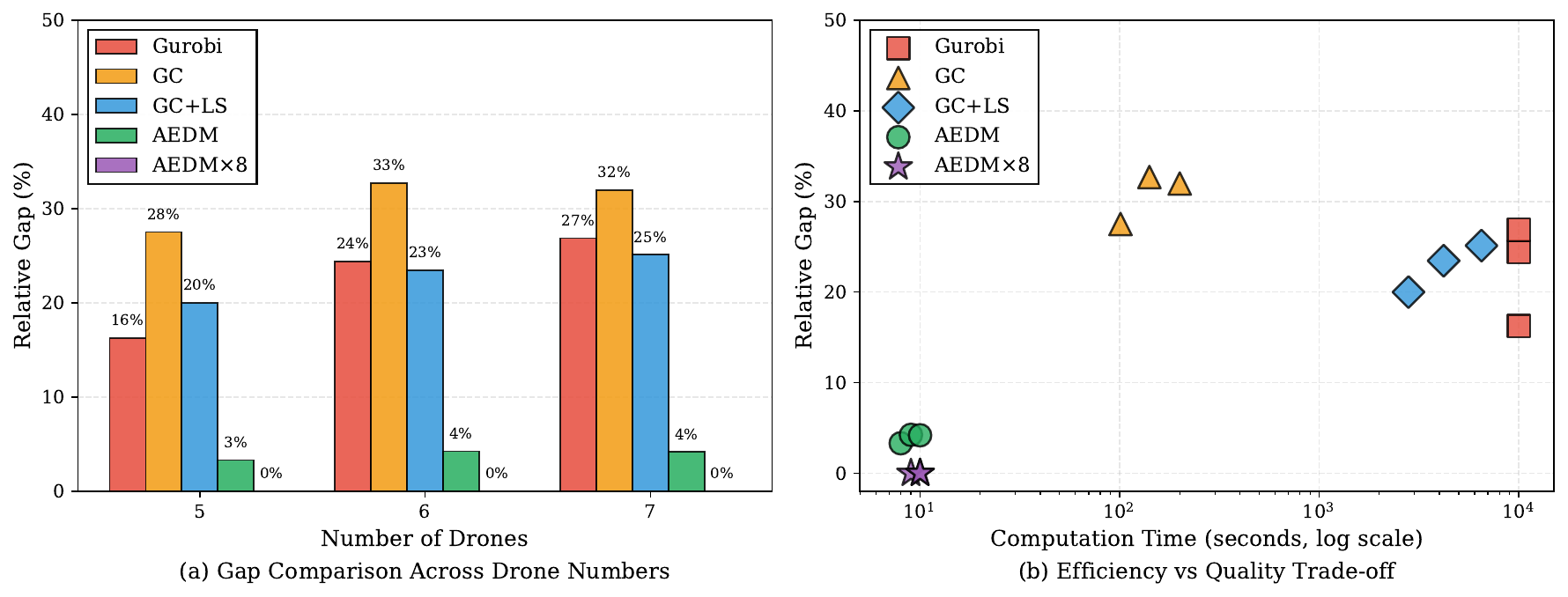}
\caption{Performance Comparison Between AEDM and Gurobi on Anaheim Road Network}
\label{fig 13}
\end{figure}

For Anaheim, we test drone numbers $K \in \{5, 6, 7\}$ with fixed $p_{\max} = 45$ minutes and Gurobi's time limit extended to 10,000 seconds due to increased problem complexity. \autoref{fig 13} presents performance comparisons across all drone configurations. \autoref{fig 13} (a) demonstrates solution quality through grouped bar charts, where AEDM×8 consistently achieves the highest objective values (115, 129, 144 for $K=5, 6, 7$), outperforming Gurobi by 16--27\%, GC by 28--32\%, and GC+LS by 20--25\%. \autoref{fig 13} (b) illustrates the efficiency-quality trade-off on logarithmic time scale, demonstrating AEDM's superiority in both solution quality and computational efficiency. AEDM achieves rapid performance with approximately 10 seconds, compared to Gurobi's full 10,000-second time limit without guarantee of optimal solutions. These results confirm AEDM's practical applicability for time-critical disaster response scenarios, where both solution quality and rapid deployment are essential.

\section{Conclusion}
\label{Conclusion}

This study addresses the challenge of rapid post-disaster road network damage assessment through drone routing optimization. We propose AEDM that determines high-quality drone assessment routes within seconds without requiring domain-specific algorithmic expertise. The key contributions include: (1) AEDM consistently outperforms both commercial solvers and traditional heuristics, achieving 20--71\% and 23--35\% improvements respectively, while maintaining rapid inference (1--2 seconds vs.\ 100--2,000 seconds); (2) A simple yet effective network transformation approach that converts a link-based routing problem into an equivalent node-based formulation; (3) A synthetic road network generation method that addresses the scarcity of large-scale training datasets for DRL applications. Experimental results demonstrate AEDM's robustness across varying problem scales, drone numbers, and time constraints, with consistent generalization to unseen parameter distributions. The model's ability to balance computational efficiency with solution quality makes it particularly suitable for time-critical disaster response scenarios where rapid decision-making is essential.

Future research can pursue two directions. First, while the proposed multi-task model addresses scenarios with multiple drone numbers and assessment time limits, it can be extended to accommodate more complex road damage assessment scenarios by integrating diverse constraints, including multiple depots, time windows for individual road links, and route ``open/closed'' assessment status. We plan to develop a unified model to tackle a broader spectrum of road damage assessment problems. Second, inspired by the scaling laws of large language models, which indicate that performance improves as parameter counts grow to the billion scale \citep{achiam2023gpt, touvron2023llama, guo2025deepseek}, we aim to investigate whether increasing the parameter count of our model can further enhance its performance.







\bibliographystyle{cas-model2-names}
\bibliography{cas-refs}

\newpage
\appendix

\section{Derivation of point $C$ coordinates}
\label{Derivation of Point $C$ Coordinates}

Given points $A(X_A,Y_A)$, $B(X_B,Y_B)$, and length $L$, we derive the coordinates $C(X_C,Y_C)$ lying on $AB$'s perpendicular bisector with $AC=L/2$ and $CB =L/2$. 

First, compute the midpoint $D$ and direction vectors:
\begin{equation}
D\left(\frac{X_A+X_B}{2}, \frac{Y_A+Y_B}{2}\right), \quad
\vec{AB} = (X_B-X_A, Y_B-Y_A), \quad
\vec{n} = (-(Y_B-Y_A), X_B-X_A) \tag{A1}
\end{equation}
The unit normal vector is:
\begin{equation}
\hat{n} = \left(\frac{-(Y_B-Y_A)}{d_{AB}}, \frac{X_B-X_A}{d_{AB}}\right), \quad
\text{where } d_{AB} = \sqrt{(X_B-X_A)^2 + (Y_B-Y_A)^2} \tag{A2}
\end{equation}

Parametrize $C$ along the bisector:
\begin{equation}
C = D + t\hat{n} \Rightarrow 
\begin{cases}
X_C = \frac{X_A+X_B}{2} + t\frac{-(Y_B-Y_A)}{d_{AB}} \\
Y_C = \frac{Y_A+Y_B}{2} + t\frac{X_B-X_A}{d_{AB}}
\end{cases} \tag{A3}
\end{equation}

Apply the length constraint $AC = L/2$:
\begin{equation}
\sqrt{\left(\frac{X_B-X_A}{2} - t\frac{Y_B-Y_A}{d_{AB}}\right)^2 + \left(\frac{Y_B-Y_A}{2} + t\frac{X_B-X_A}{d_{AB}}\right)^2} = \frac{L}{2} \tag{A4}
\end{equation}

Squaring and simplifying:
\begin{equation}
\frac{d_{AB}^2}{4} + t^2 = \frac{L^2}{4} \Rightarrow
t = \pm\frac{\sqrt{L^2-d_{AB}^2}}{2} \tag{A5}
\end{equation}

Substituting back yields the final coordinates:
\begin{equation}
X_C = \frac{X_A+X_B}{2} \mp \frac{\sqrt{L^2-d_{AB}^2}}{2}\frac{Y_B-Y_A}{d_{AB}} \tag{A6}
\end{equation}
\begin{equation}
Y_C = \frac{Y_A+Y_B}{2} \pm \frac{\sqrt{L^2-d_{AB}^2}}{2}\frac{X_B-X_A}{d_{AB}} \tag{A7}
\end{equation}

\section{Mathematical formulation}  
\label{Mathematical formulation}  

To formulate the problem, we define the binary decision variable $x_{ij}^k$, where $x_{ij}^k = 1$ if drone $k$ traverses road link $(i,j)$ for assessment, and $0$ otherwise. Additionally, we introduce the continuous flow variable $f_{ij}^k \geq 0$ representing the flow carried by drone $k$ on link $(i,j)$. Following the notation and network transformation introduced in Sections \ref{Problem definition} and \ref{Network transformation}, the drone-based road network assessment model is:

\begin{align}
\centering
\label{equation 2.3 oj}
& \max \quad \sum_{k \in K} \sum_{p \in \mathcal{P}} c_p \left( \sum_{j:(p,j) \in \bar{A}} x_{pj}^k \right)  \tag{B1} \\
\textbf{s.t.} \quad
\label{equation 2.3 link_1}
& \sum_{k \in K} \sum_{j:(p,j) \in \bar{A}} x_{pj}^k \leq 1 \quad \forall p \in \mathcal{P} \tag{B2} \\
\label{equation 2.3 balance}
& \sum_{j \in \bar{N}} x_{ji}^k  = \sum_{j \in \bar{N}} x_{ij}^k \quad  \forall i \in \bar{N}, \forall k \in K \tag{B3} \\
\label{equation 2.3 origin}
& \sum_{j \in \bar{N}} x_{oj}^k  = \sum_{i \in \bar{N}} x_{io}^k  = 1  \quad  \forall k \in K \tag{B4} \\
\label{equation 2.3 artificial_bidirection}
& x_{pj}^k + x_{jp}^k \leq 1 \quad \forall j \in \bar{N}, \forall p \in \mathcal{P}, \forall k \in K \tag{B5} \\
\label{equation 2.3 flytime}
& \sum_{(i,j) \in \bar{A}} t_{ij} x_{ij}^k \leq Q \quad \forall k \in K \tag{B6} \\
\label{equation 2.3 timelimit}
& \max_{k \in K} \left\{ \sum_{(i,j) \in \bar{A}} t_{ij} x_{ij}^k \right\} \leq p_{\max} \tag{B7}
\end{align}

\begin{itemize}
    \item Equation \eqref{equation 2.3 oj} maximizes the total information value collected by all drones, summing $c_p$ for each transformed artificial node traversed by drone $k$.
    \item Constraints \eqref{equation 2.3 link_1} restrict each information-bearing artificial node to at most one drone visit, preventing redundant assessments by avoiding overlapping visits from multiple drones. Notably, this constraint does not impose a single-visit requirement on nodes within the original road network $G$, as illustrated in \autoref{fig 3}.
    \item Constraints \eqref{equation 2.3 balance} enforce flow conservation for each node $i$ and drone $k$, ensuring equal entry and exit counts to maintain path continuity.
    \item Constraints \eqref{equation 2.3 origin} require each drone $k$ to start and end at depot $o$, with exactly one exit from and one entry to $o$.
    \item Constraints \eqref{equation 2.3 artificial_bidirection} forbid a drone from simultaneously using both links $(p,j)$ and $(j,p)$ between an artificial node $p$ and an adjacent node $j$. Since each artificial node $p \in \mathcal{P}$ represents a road link in the original network, a feasible traversal should enter $p$ from one junction and leave towards another junction. Such a pattern is an artifact of the link-to-node transformation. It does not correspond to any meaningful movement in the original road network and is therefore excluded.
    \item Constraints \eqref{equation 2.3 flytime} cap drone $k$'s total flight time at battery limit $Q$.
    \item Constraint \eqref{equation 2.3 timelimit} ensures the maximum flight time across all drones does not exceed $p_{\max}$, enforcing timely completion of the entire assessment.
\end{itemize}

Since the transformed network allows revisiting original nodes while requiring all assessed links to form connected routes from the depot, classical Miller–Tucker–Zemlin (MTZ) subtour elimination constraints used in the existing OP literature \citep{kobeaga2018efficient,zhang2023robust} are unsuitable, because they eliminate all cycles, including those that naturally occur on original nodes (see \autoref{fig 5}). To simultaneously allow such revisits and prevent disconnected subtours, we propose a single-commodity flow formulation. In this formulation, (i) the depot acts as the unique supply node that provides a total amount of flow equal to the number of artificial nodes visited by a drone; (ii) each visited artificial node consumes exactly one unit of flow, representing the completion of a damage-assessment task; and (iii) original nodes serve as pure transshipment nodes with zero net flow, allowing drones to revisit them without violating flow balance. The following formulation (B8)--(B11) eliminates disconnected subtours while permitting loops at original nodes through the differential treatment of artificial and original nodes in flow conservation constraints, addressing the key modeling challenge that traditional MTZ constraints cannot accommodate.

\begin{align}
\label{equation 2.3 flow_depot}
& \sum_{j:(o,j) \in \bar{A}} f_{oj}^k - \sum_{j:(j,o) \in \bar{A}} f_{jo}^k = \sum_{p \in \mathcal{P}} \sum_{j:(p,j) \in \bar{A}} x_{pj}^k \quad \forall k \in K \tag{B8} \\
\label{equation 2.3 flow_artificial}
& \sum_{j:(j,i) \in \bar{A}} f_{ji}^k - \sum_{j:(i,j) \in \bar{A}} f_{ij}^k = \sum_{j:(i,j) \in \bar{A}} x_{ij}^k \quad \forall i \in \mathcal{P}, \forall k \in K \tag{B9} \\
\label{equation 2.3 flow_original}
& \sum_{j:(j,i) \in \bar{A}} f_{ji}^k - \sum_{j:(i,j) \in \bar{A}} f_{ij}^k = 0 \quad \forall i \in N \setminus \{o\}, \forall k \in K \tag{B10} \\
\label{equation 2.3 flow_capacity}
& f_{ij}^k \leq |\mathcal{P}| \cdot x_{ij}^k \quad \forall (i,j) \in \bar{A}, \forall k \in K \tag{B11} \\
\label{equation 2.3 decisonlimit1}
& x_{ij}^k \in \{0,1\} \quad  \forall (i,j) \in \bar{A}, \forall k \in K  \tag{B12} \\
\label{equation 2.3 decisonlimit2}
& f_{ij}^k \geq 0 \quad \forall (i,j) \in \bar{A}, \forall k \in K \tag{B13}
\end{align}

\begin{itemize}
    \item Constraint \eqref{equation 2.3 flow_depot} establishes the depot as a supply node, with net outflow equal to the number of artificial nodes visited by drone $k$.
    \item Constraint \eqref{equation 2.3 flow_artificial} ensures each visited artificial node consumes one unit of flow, where inflow exceeds outflow by exactly one unit, representing the damage assessment task completion.
    \item Constraint \eqref{equation 2.3 flow_original} enforces flow conservation at original nodes, where inflow equals outflow, enabling node revisits while maintaining path connectivity. This is the key mechanism that allows drones to revisit original nodes as shown in \autoref{fig 5}.
    \item Constraint \eqref{equation 2.3 flow_capacity} links flow to link usage, ensuring flow is only allowed on links selected by the routing decision.
    \item Equations \eqref{equation 2.3 decisonlimit1} and \eqref{equation 2.3 decisonlimit2} define the domains of decision variables: $x_{ij}^k \in \{0,1\}$ is a binary indicator for whether drone $k$ traverses link $(i,j)$, and $f_{ij}^k \geq 0$ is the continuous flow variable on link $(i,j)$ for drone $k$.
\end{itemize}

\section{Traditional optimization methods}
\label{Traditional optimization methods}

In Appendix~\ref{Mathematical formulation}, we formulated a mathematical model for the drone-based road damage assessment problem, which can be solved using commercial solvers such as Gurobi. To enable a more detailed performance comparison and demonstrate the advantages of the proposed AEDM over traditional optimization methods, we designed a two-phase heuristic algorithm based on established heuristic frameworks for the orienteering problem (OP) \citep{kobeaga2018efficient, yang2025heuragenix}. This heuristic consists of a greedy construction phase followed by a local search improvement phase. The construction phase uses a classic profit-to-time ratio rule to generate initial solutions, while the improvement phase applies relocate, exchange, and remove-insert operators that are carefully adapted to the link-to-node transformed network structure. Unlike the proposed AEDM, however, this heuristic requires hand-crafted domain knowledge, illustrating the typical limitations of traditional algorithmic approaches in newly emerging problem settings.

The construction phase builds an initial feasible solution by sequentially assigning routes to each of the $K$ drones. For each drone $k$, starting from the depot $o$, the algorithm iteratively selects the next artificial node $p \in \mathcal{P} \setminus \mathcal{V}$ (where $\mathcal{V}$ denotes the set of already visited artificial nodes) that maximizes the profit-to-time ratio:
\begin{equation}
\text{score}(p) = \frac{c_p}{\text{ShortestPathTime}(u \to p)}
\tag{C1}
\end{equation}
where $u$ is the current node and $c_p$ is the information value of artificial node $p$. The selection is subject to the following constraints: (i) a connectivity constraint requiring that the path $u \to p$ be feasible in the transformed network $\bar{G}$, ensuring that the next node is reachable from the current position; (ii) time feasibility, requiring that the accumulated flight time plus the time to reach $p$ and return to the depot does not exceed the time limit $\min\{Q, p_{\max}\}$; and (iii) exclusivity constraint $p \notin \mathcal{V}$, preventing redundant assessment of the same road link. Once a node is selected, the shortest path from $u$ to $p$ is computed and appended to the current drone's route. This process continues until no more feasible nodes can be added, at which point the drone returns to the depot and the next drone begins its route construction. The greedy construction phase terminates when all $K$ drones have been assigned routes or all artificial nodes with positive information values have been visited.

After obtaining an initial solution from construction phase, the local search phase attempts to improve solution quality through iterative neighborhood exploration. Given the unique structure of the transformed network, we adapt three local search operators to work with triplet structures. A triplet $(u_p, p, v_p)$ represents an artificial node $p$ along with its predecessor $u_p$ and successor $v_p$ in a drone's route. The three operators are defined as follows:

\begin{itemize}
    \item Relocate: Reverse the direction of a triplet within the same route by changing $(u_p, p, v_p)$ to $(v_p, p, u_p)$. This operator explores alternative approach directions to the same road link.
    
    \item Remove-insert: Remove a triplet $(u_p, p, v_p)$ from its route and insert a different triplet $(u_q, q, v_q)$ at a new position. This operator removes an existing assessment task and replaces it with an alternative one, potentially improving the route's total information value.
    
    \item Exchange: Swap two triplets $(u_p, p, v_p)$ and $(u_q, q, v_q)$ between two different drone routes, potentially improving the overall solution by reassigning assessment tasks.
\end{itemize}

The local search procedure iteratively evaluates all possible moves using these three operators. At each iteration, if a feasible neighboring solution with higher total information value is found, the current solution is updated. The search continues for a maximum of $\text{max\_iterations}=1,000$ iterations. The final solution $\Pi^*$ with the highest reward $R^*$ encountered during the search is returned as the output of the heuristic algorithm.

\begin{algorithm}[!ht]
\caption{Greedy Heuristic with Local Search}
\label{algorithm2}
\begin{algorithmic}[1]

\STATE \textbf{Input:} Transformed network $\bar{G} = (\bar{N}, \bar{A})$, flight times $t_{ij}$, information values $c_p$ for $p \in \mathcal{P}$, depot $o$, number of drones $K$, battery flight time limit $Q$, maximum allowable assessment time $p_{\max}$, maximum local-search iterations $max\_iterations$

\vspace{0.5em}
\STATE \textbf{Phase 1: Greedy construction of initial solution}
\STATE Build adjacency structure from $\bar{A}$
\STATE Initialize visited artificial nodes: $\mathcal{V} \leftarrow \emptyset$
\STATE Initialize solution: $\Pi \leftarrow \emptyset$, total reward $R \leftarrow 0$
\STATE Initialize global time limit $T_{\max} = \min\{Q,\, p_{\max}\}$

\FOR{$k = 1$ to $K$}
    \STATE Construct greedy route $\pi^k$:
    \STATE \quad Set current node $u \leftarrow o$
    \STATE \quad Iteratively select next artificial node $p \in \mathcal{P} \setminus \mathcal{V}$ maximizing
    \STATE \qquad 
        $\displaystyle \text{score}(p) = 
        \frac{c_p}{\text{ShortestPathTime}(u \rightarrow p)}$
    \STATE \quad Subject to:
            (i) feasibility of the path $u \to p$ in the transformed network $\bar{G}$,
           (ii) accumulated time $\le T_{\max}$,
           (iii) feasibility of returning to $d$,
           (iv) $p \notin \mathcal{V}$
    \STATE \quad Append shortest-path segments to construct complete route $\pi^k$
    \STATE \quad $\mathcal{V} \leftarrow \mathcal{V} \cup \{p\}$
    \STATE Add $\pi^k$ into $\Pi$; update $R$
    
\ENDFOR

\vspace{0.5em}
\STATE \textbf{Phase 2: Local search improvement}
\STATE Initialize best solution: $\Pi^* \leftarrow \Pi$, $R^* \leftarrow R$
\STATE \quad Define local operators: 
\STATE (a) Relocate:
Change a triplet $(u_p,p,v_p)$ into $(v_p,p,u_p)$ within the same route.
\STATE (b) Remove-insert:
Remove triplet $(u_p,p,v_p)$ from its route and insert $(u_q,q,v_q)$ at a new position.
\STATE (c) Exchange:
Swap two triplets $(u_p,p,v_p)$ and $(u_q,q,v_q)$ between routes.
\FOR{$iteration = 1$ to $max\_iterations$}
    \FOR{operator $\in$ \{Relocate, Remove-insert, Exchange\}}
        \STATE Generate candidate solution $\Pi'$ using operator
        \IF{$\Pi'$ feasible \textbf{and} $R(\Pi') > R(\Pi)$}
            \STATE $\Pi \leftarrow \Pi'$
            \IF{$R(\Pi') > R^*$}
                \STATE $\Pi^* \leftarrow \Pi'$, \; $R^* \leftarrow R(\Pi')$
            \ENDIF
            \STATE \textbf{break}
        \ENDIF
    \ENDFOR

\ENDFOR

\STATE \textbf{Output:} Optimized routes $\Pi^*$ and reward $R^*$

\end{algorithmic}
\end{algorithm}

The complete pseudocode for the two-phase heuristic is provided in Algorithm \ref{algorithm2}. It is important to note that while this heuristic can produce reasonable solutions, its design required substantial domain expertise to: (1) adapt the profit-to-time ratio heuristic to the transformed network structure, (2) develop triplet-based operators that respect the exclusivity constraints of artificial nodes while allowing revisits to original nodes, and (3) carefully tune the search parameters and stopping criteria. This design complexity highlights a key advantage of the proposed AEDM: it eliminates the need for such hand-crafted problem-specific heuristics by learning effective routing policies directly from data.

\section{Ablation study} 
\label{Ablation study}

To evaluate the impact of the proposed reward normalization strategy (EMA with Z-score normalization) on model performance, we conducted an ablation study by comparing four normalization variants of the AEDM framework:
\begin{enumerate}[label=\alph*)]
    \item EMA with Z-score normalization: Combines EMA of both mean and variance to compute Z-score normalized rewards, as proposed in \eqref{15}-\eqref{17}.
    \item Batch-wise Z-score normalization: Normalizes rewards within each batch using the batch mean and variance, without exponential smoothing.
    \item EMA mean division normalization: Normalizes rewards by dividing by the EMA of the mean (ignoring variance).
    \item No normalization: Uses raw rewards without any regularization.
\end{enumerate}

\begin{figure}
\centering
\includegraphics[width=0.7\textwidth]{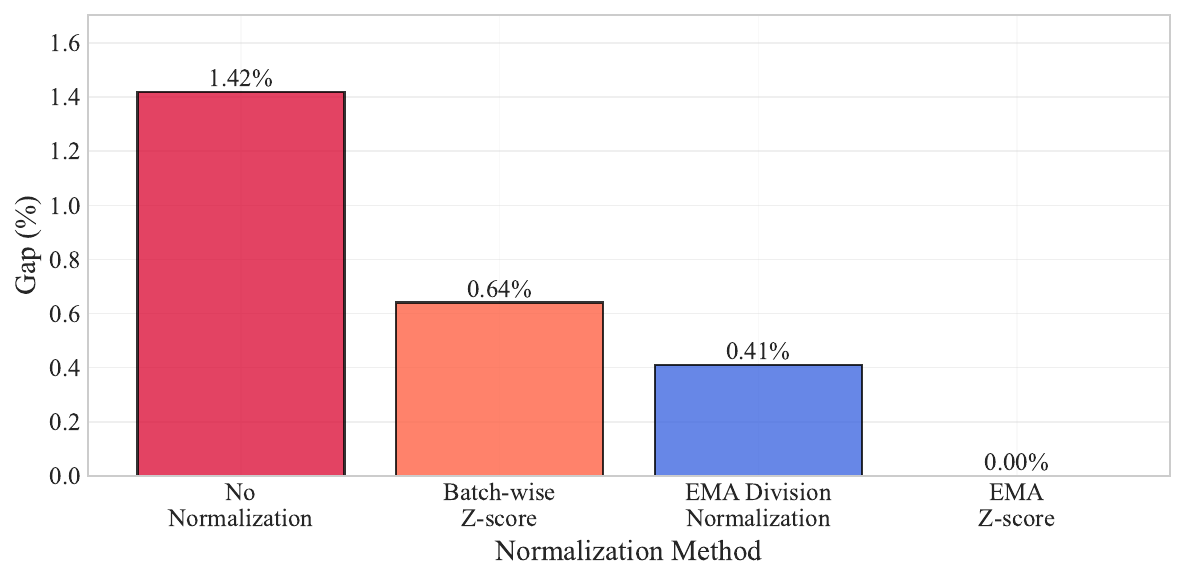}
\caption{Performance Gap of AEDM Under Diverse Reward Normalization Methods}
\label{fig 14}
\end{figure}

All variants were trained under identical conditions detailed in Section~\ref{Experiment setting} and evaluated on 400-node networks with $p_{\text{max}} = 45$ minutes and 4, 5, 6, 7 drones (i.e., four scenarios). Performance was quantified by the average relative gap across these four scenarios, defined as $\text{Gap} = (y-y_{\text{other}})/y$. Here, $y$ denotes the objective value from the EMA with Z-score normalization, and $y_{\text{other}}$ represents that from other normalization variants. As visualized in \autoref{fig 14}, the proposed reward normalization strategy achieves the best performance, whereas the “No normalization” variant yields the worst. These results underscore the critical role of EMA-based mean and variance tracking in stabilizing reinforcement learning for multi-task drone routing. The proposed normalization strategy ensures reward comparability across diverse parameter regimes (e.g., varying drone numbers and time limits), enabling effective model generalization to unseen scenarios. 

\section{Method interpretability} 
\label{Method interpretability}

\begin{figure}
\centering
\includegraphics[width=0.98\textwidth]{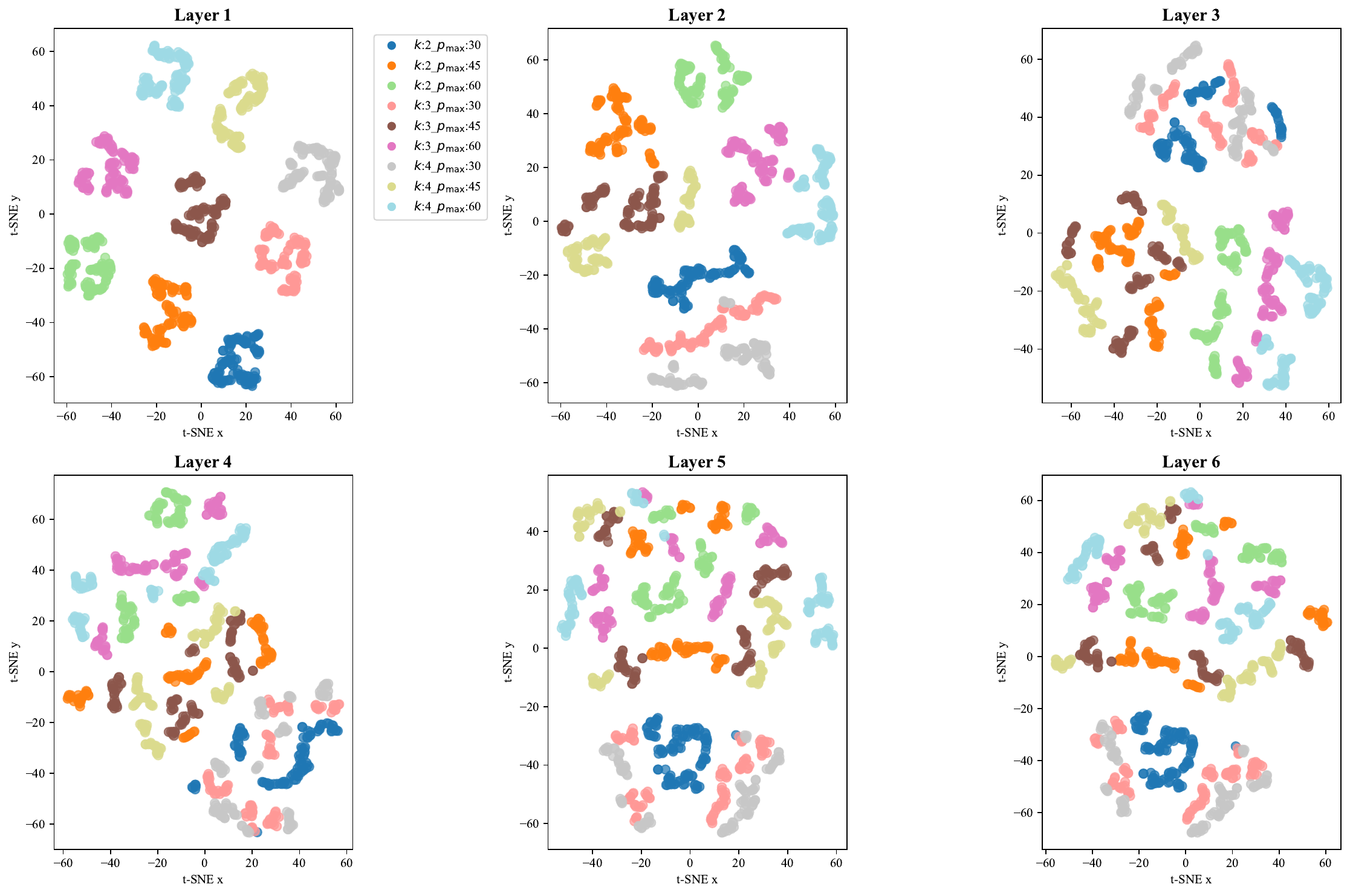}
\caption{Method Interpretability: Encoder Latent Space via Layer-Wise t-SNE Analysis}
\label{fig 15}
\end{figure}

To explore the model's interpretability, we employ t-distributed stochastic neighbor embedding (t-SNE) to project high-dimensional representations, which are learned under varying drone numbers $k$, $p_{\max}$, and a fixed 100-node scale, into a 2D space. With 100 instances per parameter combination, \autoref{fig 15} visualizes the evolution of features across the output of each encoder layer.

In Layer 1, features cluster distinctly by $k$-$p_{\max}$ combinations, indicating that the model first learns superficial parameter distinctions in initial layers and validating its ability to recognize input configurations. Progressing to Layer 2, clusters remain distinct but exhibit early fusion, where features begin interacting across $k$ and $p_{\max}$ while retaining core differences. In Layers 3 and 4, clusters further soften and then mix: distributions of different $k$/$p_{\max}$ values interweave, reducing strict separation. In Layers 5 and 6, clusters dissolve into complex, interwoven distributions. This layer-by-layer t-SNE evolution reveals a clear learning trajectory: initial layers (1--2) recognize and separate raw input parameters ($k$, $p_{\max}$); middle layers (3--4) fuse these features; final layers (5--6) abstract core problem structures, prioritizing solution logics over input specifics. This confirms the model learns generalizable principles (not merely parameter patterns), leveraging the transformer architecture to distill raw inputs into abstract representations of the core problem. The cross-layer feature fusion validates that the model transitions from "memorizing" to "reasoning", which is a critical hallmark of effective problem-solving in complex tasks.

\end{document}